\documentclass[10pt,twocolumn,letterpaper]{article}

\usepackage[pagenumbers]{cvpr} 
\usepackage{graphicx}
\usepackage{amsmath}
\usepackage{amssymb}
\usepackage{booktabs}

\usepackage{tabu}
\usepackage{multirow}
\usepackage{rotating} 
\usepackage{caption}
\usepackage{subcaption}
\usepackage{pifont}
\newcommand{\cmark}{\checkmark}%
\newcommand{\xmark}{$\times$}%
\newcommand{\good}[1]{\fontsize{6}{6}\selectfont{\textcolor{blue}{(+#1)}}}
\newcommand{\bad}[1]{\fontsize{6}{6}\selectfont{\textcolor{red}{(-#1)}}}

\newcommand{\goodi}[1]{\fontsize{5}{6}\selectfont{\textcolor{blue}{(+#1)}}}
\newcommand{\badi}[1]{\fontsize{5}{6}\selectfont{\textcolor{red}{(-#1)}}}

\newcommand{\Sref}[1]{Sec.~\ref{#1}}
\newcommand{\Eref}[1]{Eq.~(\ref{#1})}
\newcommand{\Fref}[1]{Fig.~\ref{#1}}
\newcommand{\Tref}[1]{Table~\ref{#1}}
\newcommand{\R}[0]{\mathbb{R}}

\newcommand{\myparagraph}[1]
{
\vspace{2mm}\noindent\textbf{#1}
}
\newcommand{\supref}[1]{{\color{red}#1}}

\newcommand {\junyan}[1]{}

\newcommand{\tb}[1]{\textbf{#1}}

\definecolor{methodred}{RGB}{248, 203, 173}
\definecolor{methodyellow}{RGB}{245, 220, 143}
\definecolor{methodgreen}{RGB}{197, 224, 180}
\definecolor{methodblue}{RGB}{180, 199, 231}

\definecolor{cvprblue}{rgb}{0.21,0.49,0.74}
\usepackage[pagebackref,breaklinks,colorlinks,citecolor=cvprblue]{hyperref}

\usepackage[capitalize]{cleveref}
\crefname{section}{Sec.}{Secs.}
\Crefname{section}{Section}{Sections}
\Crefname{table}{Table}{Tables}
\crefname{table}{Tab.}{Tabs.}
\def\paperID{7637} 
\def\confName{CVPR}
\def\confYear{2024}

\title{Grounded Text-to-Image Synthesis with Attention Refocusing}

\author{Quynh Phung~~~~Songwei Ge~~~~Jia-Bin Huang\\
    University of Maryland College Park\\
\url{https://attention-refocusing.github.io/}
}

\begin{document}
\twocolumn[{%
\renewcommand\twocolumn[1][]{#1}%
\maketitle
\vspace{-3em}
\begin{center}
    \centering
    \captionsetup{type=figure}
  
    \begin{subfigure}[t]{1.\textwidth}
   
      \includegraphics[width=1.\textwidth]{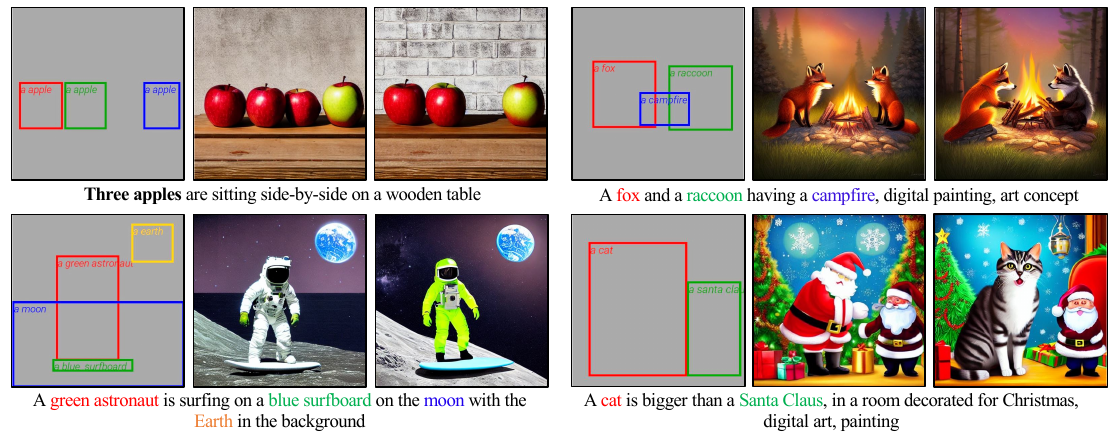}
    \vskip -0.cm
    {\small
      \begin{tabu} to 1.\textwidth {X[2cm,c]X[2cm,c]X[2.2cm,c]X[2.2cm,c]X[2cm,c]X[2cm,c]}
        Input & GLIGEN ~\cite{li2023gligen} &GLIGEN + Ours &  Input & GLIGEN ~\cite{li2023gligen} & GLIGEN + Ours \\
        \end{tabu}}
    \end{subfigure}
     \vskip 0.3cm

    \begin{subfigure}[t]{1.\textwidth}
    
      \includegraphics[width=1.\textwidth]{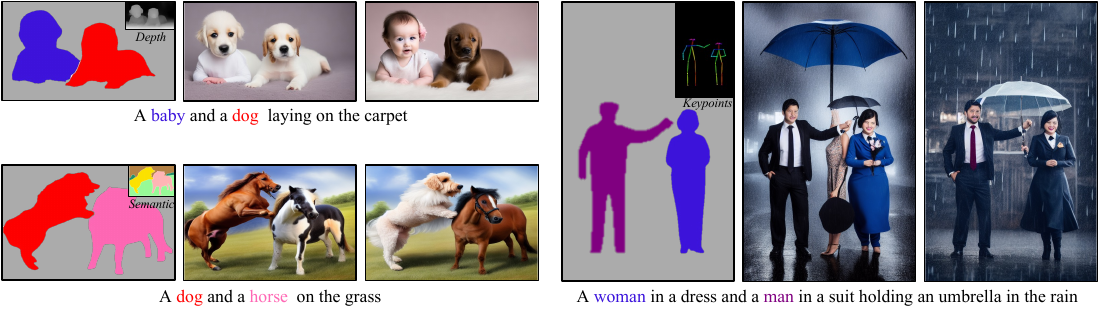}
      \vskip -0.1cm
    {\small
      \begin{tabu} to 1.\textwidth {X[2cm,c]X[2cm,c]X[2.2cm,c]X[2.2cm,c]X[2cm,c]X[2cm,c]}
        Input & ControlNet ~\cite{zhang2023adding} &ControlNet + Ours &  Input & ControlNet ~\cite{zhang2023adding} & ControlNet + Ours \\
        \end{tabu}}
      
    \end{subfigure}
    
    \vskip -0.1cm
    \setcounter{figure}{0}
    \captionof{figure}{
    \tb{Controllable text-to-image synthesis with attention refocusing.}
We introduce a new framework to improve the controllability of text-to-image synthesis. 
Given a text prompt, e first leverage GPT-4 to generate spatial layouts and then use grounded text-to-image methods to generate the images given the layouts and prompts. 
However, the detailed information, like the quantity, identity, and attributes, often remains incorrect or mixed using existing models. 
We propose a training-free approach --- attention-refocusing --- to substantially improve the controllability. 
Our method is model-agnostic and can be applied to enhance the control capacity of methods like GLIGEN ~\cite{li2023gligen} and ControlNet~\cite{zhang2023adding}
}
    
    \label{fig:teaser}
\end{center}%
}]

\begin{abstract}

Driven by the scalable diffusion models trained on large-scale datasets, text-to-image synthesis methods have shown compelling results. 
However, these models still fail to precisely follow the text prompt involving multiple objects, attributes, or spatial compositions.
In this paper, we reveal the potential causes in the diffusion model's cross-attention and self-attention layers. 
We propose two novel losses to refocus attention maps according to a given spatial layout during sampling. 
Creating the layouts manually requires additional effort and can be tedious.
Therefore, we explore using large language models (LLM) to produce these layouts for our method.
We conduct extensive experiments on the DrawBench, HRS, and TIFA benchmarks to evaluate our proposed method. 
We show that our proposed attention refocusing effectively improves the controllability of existing approaches. 
\vspace{-0.5cm}
\end{abstract}

\vspace{-0.2cm}
\section{Introduction}
\label{sec:introduction}

Despite the unprecedented zero-shot capacity and photorealism achieved by the recent progress in text-to-image synthesis~\cite{DALLE, Imagen, balaji2022ediffi, Parti, DALLE2, rombach2022high, kang2023gigagan}, existing models still struggle with text prompts containing multiple objects and attributes with complex spatial relationships among them~\cite{chefer2023attend,bakr2023hrs,chen2023training,feng2022training,xiao2023fastcomposer}. 
Some objects, attributes, and spatial compositions specified in the text prompts are often mixed, swapped, or even completely missing in the synthesized image.
Our work aims to mitigate this problem by grounding the text-to-image synthesis using explicit spatial layouts \emph{without extra training}.

The deep level of language understanding exhibited by the text-to-image models can be attributed to using pretrained language models~\cite{radford2021learning} as the text encoder~\cite{Imagen}. 
The computed text embeddings are processed using the \emph{cross-attention layers} in the denoising models~\cite{GLIDE,nichol2021improved}.
Upon careful analysis of the failure examples generated by Stable Diffusion~\cite{rombach2022high}, we identify a potential cause of the failure above in the attention layers~\cite{vaswani2017attention}, where the pixels with similar features produce similar attention queries and consequently attend to a similar set of regions or tokens. 
The information of these pixels is thus \emph{mixed} through these attention layers. 
Note that such pixels can come from two different objects with similar features. 
For example, given the prompt ``A dog on the right of a cat'', a pixel associated with the token ``dog'' could have similar features to the pixels in the ``cat'' region.
As a result, the model could incorrectly attend to the ``cat'' token through the cross-attention layers or the ``cat'' region through self-attention layers, causing the missing object or blended attribute issues.

Previous studies propose to mitigate this issue by manipulating the cross-attention maps during the sampling process~\cite{feng2022training, chen2023training, chefer2023attend}. 
However, they overlook a similar issue in self-attention layers, where distinguishing between pixels of the same object and those of different objects with similar features becomes a challenge. To this end, we leverage \emph{explicit layout representations} for grounded synthesis following the previous works~\cite{li2023gligen,chen2023training}.


 
In this paper, we propose two novel losses based on the input layout during the sampling process to \emph{refocus} the attention in both self- and cross-attention layers. 
Our analysis shows that with our losses the attention can be effectively \emph{refocused} to the desired region instead of similar but irrelevant regions.
We also explore using LLMs to generate explicit layout representations. 
We demonstrate that these models have strong spatial reasoning capabilities and can predict the plausible layout of the objects when using our designed prompts with in-context learning.
We will release code and data in the future.

We show that when combining the bounding boxes generated by GPT4~\cite{openai2023gpt4} and our attention-refocusing losses, our method significantly and consistently improves over several strong baselines on the DrawBench~\cite{Imagen}, HRS benchmarks~\cite{bakr2023hrs}, and TIFA benchmark~\cite{hu2023tifa}. 
Our main contributions are summarized below:
\begin{itemize}
    \item We propose attention-refocusing losses to regularize both cross- and self-attention layers during the sampling to improve the controllability given the layout and text prompt;
    \item We explore using LLMs to generate layouts given text prompts, allowing the exploitation of the up-to-date LLMs with trained text-to-image models;
    \item We conduct a comprehensive experiment on existing methods of grounded text-to-image generation and show that our method compares favorably against the state-of-the-art models.
\end{itemize}
\vspace{-0.3cm}

\section{Related work}
\vspace{-0.3cm}






\begin{figure*}[t]
\centering
\includegraphics[width=\textwidth]{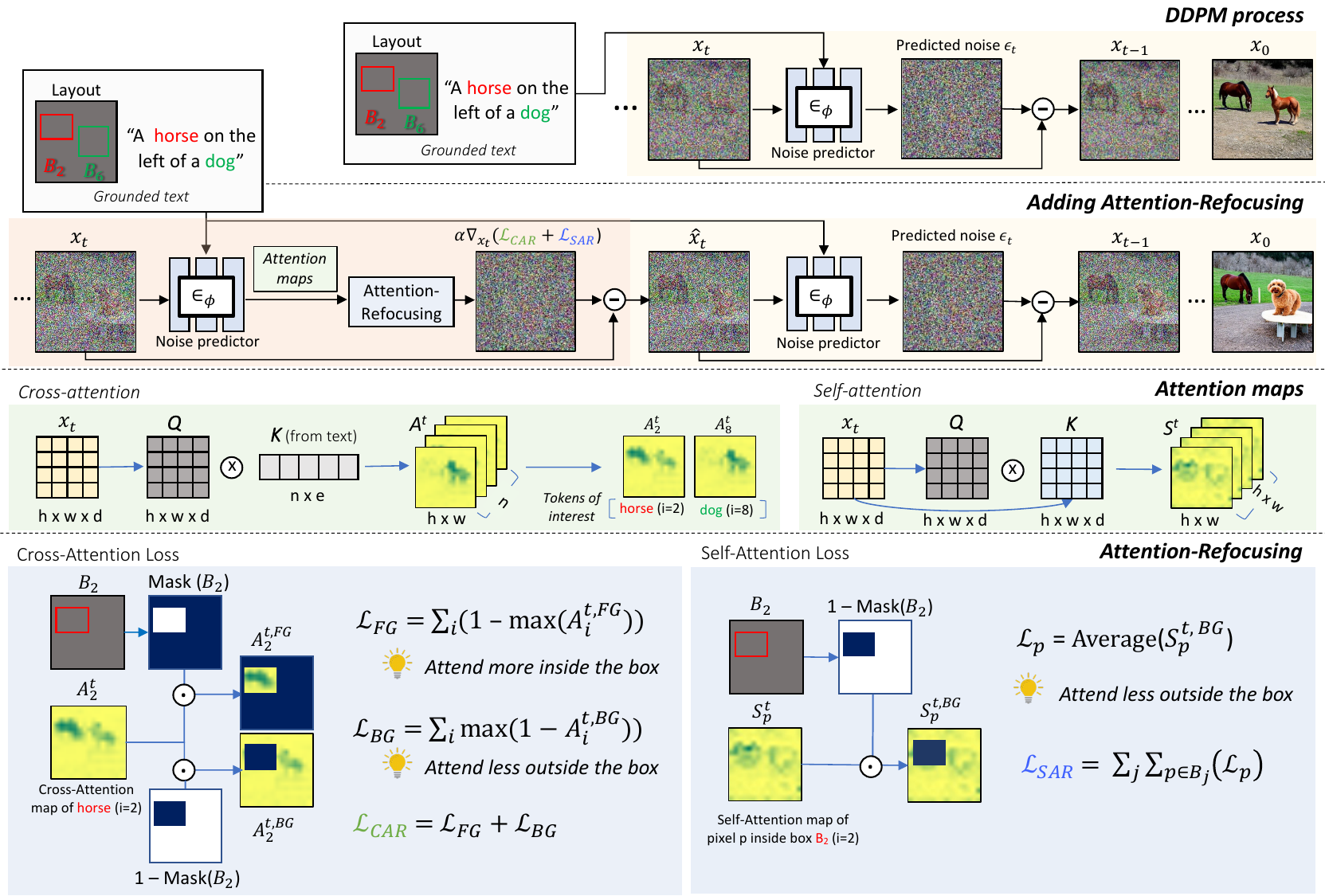}
\caption{
\tb{The proposed Attention-Refocusing framework.}
At each denoising step, we update the noised sample 
by optimizing our $\mathcal{L}_{CAR}$ and $\mathcal{L}_{SAR}$ losses (\textcolor{methodred}{red} block) before denoising with the predicted noise  (\textcolor{methodyellow}{yellow} block).
For each cross-attention map, $\mathcal{L}_{CAR}$ is designed to encourage a region to attend more to the corresponding token while discouraging the remaining region from attending to that token. For each self-attention map, $\mathcal{L}_{SAR}$ prevents the pixels in a region from attending to irrelevant regions ($\mathcal{L}_{CAR}$ and $\mathcal{L}_{SAR}$  in \textcolor{methodblue}{blue}  blocks). 
}
\label{fig:framework}
\vspace{-0.6cm}
\end{figure*}

\myparagraph{Large-scale text-to-image models} 
High-resolution text-to-image synthesis has been dramatically advanced by the development of large-scale text-to-image models~\cite{DALLE,Imagen,Parti,balaji2022ediffi,rombach2022high,kang2023gigagan,Make-A-Scene}. 
Such rapid progress can be attributed to several critical techniques. 
First, the availability of large-scale text-image datasets~\cite{schuhmann2022laion,kakaobrain2022coyo-700m} makes it possible to train data-hungry models on a massive volume of samples from diverse resources. 
The development of the scalable model architectures, including GANs~\cite{sauer2023stylegan,kang2023gigagan}, autoregressive models~\cite{DALLE,Parti,chang2023muse,ding2022cogview2}, and diffusion models~\cite{ho2020denoising,balaji2022ediffi,DALLE2,GLIDE,Imagen}, together with various training and inference techniques~\cite{ho2020denoising, song2021denoising,ho2022cascaded,Classifier}. 
Our work focuses on the problem of improving the \emph{controllability} of the generated images with respect to the input text with large-scale diffusion models. 

\myparagraph{Improving the controllability of text-to-image models}
Enhancing the user control of large-scale text-to-image models has drawn great attention recently. 
Previous work proposes to boost the controllability through various input formats such as rich text~\cite{ge2023expressive}, personal images~\cite{ruiz2022dreambooth,kumari2022multi}, edge maps, segmentation masks, depth maps ~\cite{zhang2023adding}, and bounding boxes~\cite{avrahami2022spatext,bar2023multidiffusion,li2023gligen}. 
There are also works focusing on strengthing the controllability of the original input text, motivated by the observation that existing models often fail to fulfill the description from the input text accurately~\cite{feng2022training,paiss2023teaching,chefer2023attend,bakr2023hrs}.
For example, when multiple objects and attributes occur in the text prompt, some are often missing or mixed in the synthesized images~\cite{feng2022training,chefer2023attend}. 
Attend-and-Excite~\cite{chefer2023attend} proposes optimizing cross-attention maps during sampling to ensure all the tokens are attended to. 
Several studies finetune the existing models with human feedback~\cite{lee2023aligning,wu2023better} or use improved language models~\cite{liu2022character,paiss2023teaching,zhong2023adapter,touvron2023llama,xue2022byt5} to enhance the text-image alignment. 
Similar to these recent efforts, our work also focuses on improving the alignment between the generated images and input texts.
However, we leverage an intermediate spatial layout generated by LLMs~\cite{radford2021learning,raffel2020exploring,openai2023gpt4,touvron2023llama} and ground the image synthesis on the layout.

\vspace{-0.2cm}

\myparagraph{Layout-conditioned text-to-image synthesis.} 
Several approaches have been developed to extend the Stable Diffusion ~\cite{rombach2022high} to condition its generation on the layouts through finetuning on layout-image pairs ~\cite{avrahami2022spatext,li2023gligen,zhang2023adding,GLIDE} or modifying the sampling process ~\cite{bar2023multidiffusion,chen2023training,balaji2022ediffi,bansal2023universal}. 
For example, GLIGEN ~\cite{li2023gligen} finetunes a gated self-attention layer to incorporate the box information from the input to the Stable Diffusion model. 
Mixture-of-Diffusion ~\cite{jimenez2023mixture} and MultiDiffusion ~\cite{kumari2022multi} perform a denoising process on each region and fuse the predicted scores. 
Others, including SD-Pww ~\cite{balaji2022ediffi}, layout-predictor ~\cite{wu2023harnessing}, direct-diffusion ~\cite{ma2023directed} Layout-guidance ~\cite{chen2023training}, and BoxDiff ~\cite{xie2023boxdiff}, directly optimize the cross-attention layers during the sampling process. However, our approach not only optimizes cross-attention maps but also self-attention maps, which is not commonly addressed by these methods. Unlike the optimization of multiple values in these methods, which can lead to image quality degradation, our method iteratively optimize peak values in the attention maps, preserving image quality. 
Universal guidance ~\cite{bansal2023universal} leverages a trained object detector and constructs a loss to force the generated images to match location guidance. 
DenseDiffusion ~\cite{kim2023dense} directly modifies all attention maps based on mask guidance without any optimization steps. Differing from DenseDiffusion, our method optimizes the latent space, indirectly influencing attention maps under mask guidance. Our approach yields improvements demonstrated in our experiments.
Our proposed method to ground the text-to-image generation on the layout uses both \emph{cross-attention} and \emph{self-attention} layers without needing extra training or additional models. 
We demonstrate that adding the proposed attention-based guidance to various base models improves their performance consistently. 

\vspace{-0.1cm}

\myparagraph{Layout predictions.}
Several concurrent works leverage the potential of large language models for enhancing text-to-image models. Similar to ours, the concurrent work LLM-grounded Diffusion ~\cite{lian2023llmgrounded} uses GPT-4 as a layout generator. Their extended work, LDM ~\cite{lian2023llm}, produces dynamic scene layouts rather than single layouts, guiding a diffusion model for video generation. 
Another concurrent work, LayoutGPT ~\cite{feng2023layoutgpt}, leverages GPT to create layouts from text conditions, then uses GLIGEN ~\cite{li2023gligen} to generate images from the created layouts. 
Cho et al. ~\cite{cho2023visual} finetune an open-source language model for the specific text-to-layout task and use standard layout-to-image models for image generation. 
However, as demonstrated in the experimental results, existing grounded text-to-image models still fail to fulfill the details in the text prompts, like quantity, identity, and attributes.
Our work further improves the controllability using attention-based guidance.
\vspace{-0.2cm}

\label{sec:relatedwork}  
\begin{figure}[t]
\vspace{-0.5em}
    \centering
    \includegraphics[width=\linewidth]{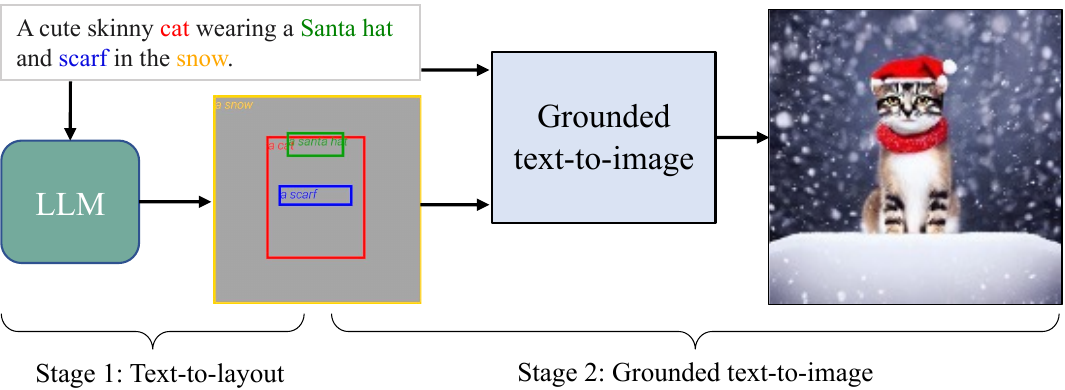}
    \caption{\tb{Our pipeline.} Our approach includes 1) text-to-layout using GPT-4 model and 2) grounded text-to-image using a pretrained diffusion model with our attention-refocusing.}
    \label{fig:overview}
    \vspace{-0.5cm}
\end{figure}

\section{Method}
\label{sec:method}
\vspace{-0.2cm}
\begin{figure*}[t]
    \centering
    \begin{subfigure}[b]{0.49\textwidth}
        \includegraphics[width=1.\textwidth]{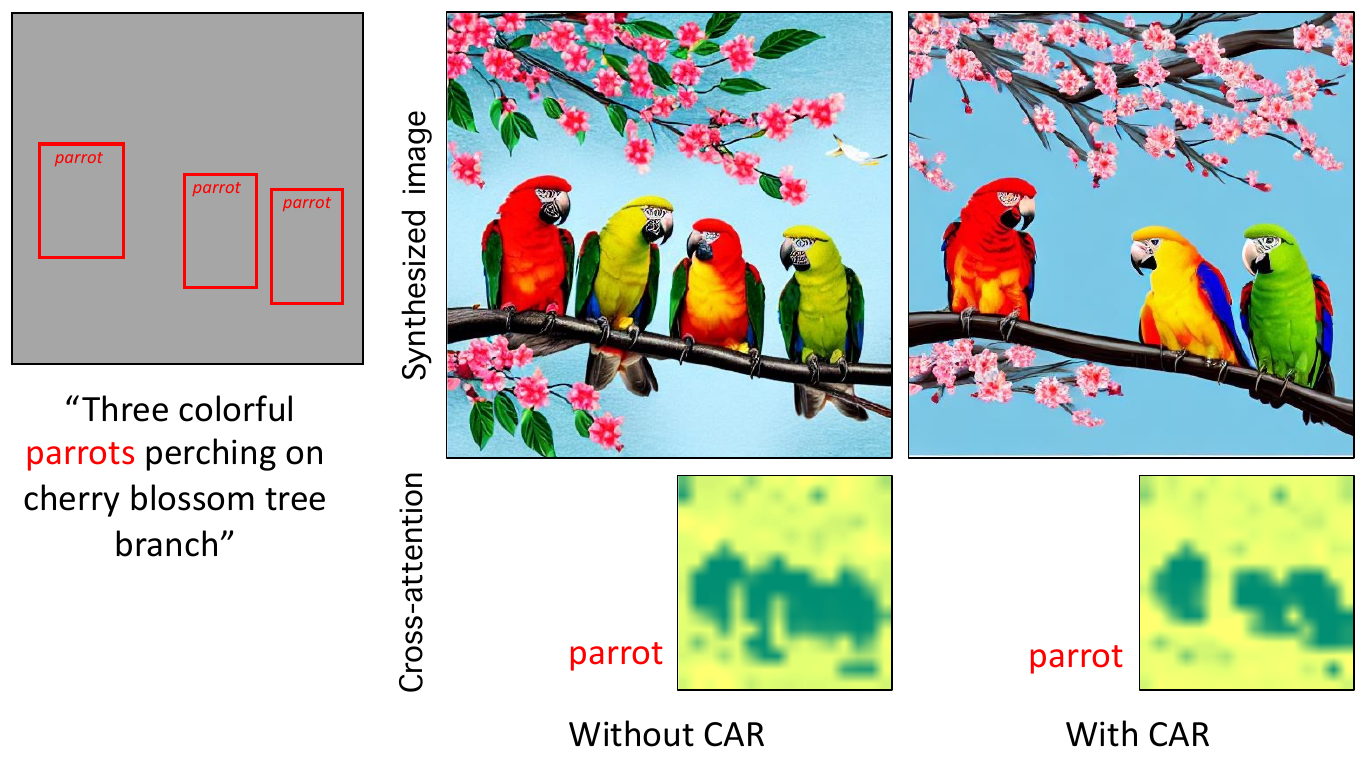}
        \vskip -0.5cm
        (a)
        \label{fig:car_motivation}
    \end{subfigure}
    \hfill
    \begin{subfigure}[b]{0.49\textwidth}
        \includegraphics[width=1.\textwidth]{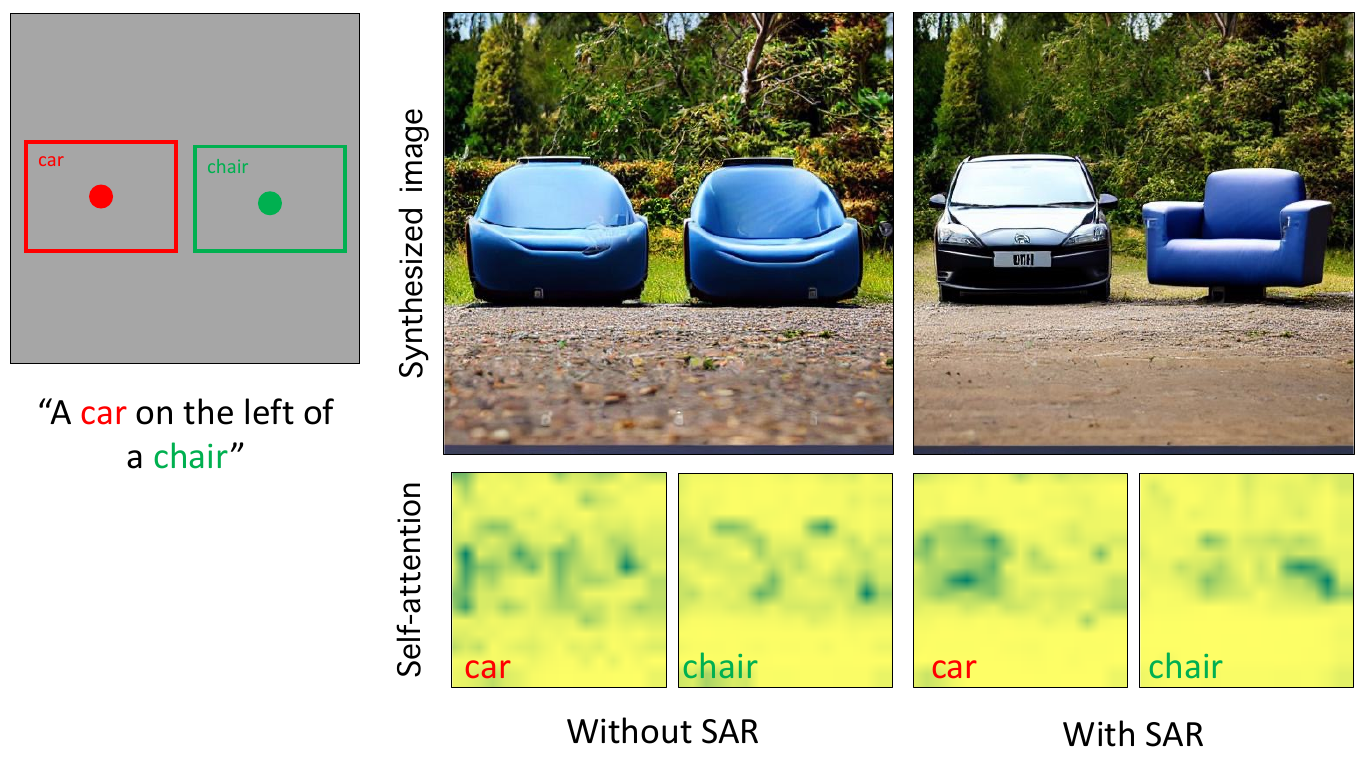}
        \vskip -0.5cm
        (b)
        \label{fig:sar_motivation}
    \end{subfigure}
    \vspace{-0.1cm}
    \caption{
(a) \tb{Cross-Attention-Refocusing (CAR) visualization.}
Without CAR, the token "parrot" attends to background regions.
Using CAR calibrates the cross-attention map to attend to the correct regions. 
(b) \tb{Self-Attention-Refocusing (SAR) visualization.} 
The dots in each box represent the pixel query of the self-attention map. 
Applying SAR loss helps refocus the self-attention layer to attend \emph{less} to the incorrect regions.
    }
    \label{fig:motivation}
    \vspace{-0.4cm}
\end{figure*}

In this section, we discuss our method for grounded text-to-image generation.
Our approach includes two main phases, as shown in~\Fref{fig:overview}: 
1) text-to-layout and 
2) grounded text-to-image. 
In both phases, we use off-the-shelf pretrained models \emph{without any extra training}. 
We exploit the spatial understanding ability in the latest large language models (LLMs) to produce visual representations such as bounding boxes as the layout given a text prompt. 

\subsection{Preliminaries}

\label{sec:method_prelim}
\vspace{-0.5em}
\myparagraph{Text-to-image diffusion models.} 
The key to the text-to-image diffusion model is the iterative denoising process. 
A UNet model is trained to progressively denoise the random Gaussian noise by computing the score $\mathbf{\epsilon}_t = U(\mathbf{x}_{t}; \mathbf{c})$, where $t$ is the time step and $\mathbf{c}$ is the embedding for conditional information. 
Next, we briefly describe the two types of attention layers used in our method.

\myparagraph{Cross-attention layer.} 
Text-to-image diffusion models condition its generation on the text prompt via cross-attention layers. 
Specifically, a pretrained CLIP encoder~\cite{radford2021learning} is often used to encode the text prompt $\mathbf{w}=(w_1, w_2, \cdots w_n)$ and obtain the text embeddings $\mathbf{c} = f_\text{CLIP}(\mathbf{w}) \in \R^{n \times e}$, where $e$ is the embedding dimension. 
The \emph{key} $\mathbf{K}\in \R^{n\times d}$ and \emph{value} $\mathbf{V}\in \R^{n\times d}$ are obtained from text embedding $c$ with a linear mapping ($d$ is the feature dimension). 
As shown in the third row of the ~\Fref{fig:framework}, given a set of queries $\mathbf{Q} \in \R^{hw \times d}$ computed from the features map of size $h \times w$, the cross-attention map $A^t$ at the step $t$ is computed as: 
\begin{equation}
 \vspace{-0.1cm}
\label{eq:attention}
    A^t = \text{softmax} \left(\frac{\mathbf{QK^\top}}{\sqrt{d}}\right) \in [0,1]^{hw\times n},
    \vspace{-0.1cm}
\end{equation}
which is formed by $n$ attention maps $\{A^t_1, ..., A^t_n\}$, where $A^t_i \in [0,1]^{h\times w}$ denotes the strength of association between a word token $w_i$ and each spatial location in the feature map.

\myparagraph{Self-attention layer.} It is used to facilitate the use of global information. 
It propagates the feature at each spatial location to a similar region in the feature map of resolution $h \times w$. 
With all key-value-query obtained from the same feature map through linear mappings and \Eref{eq:attention}, the self-attention map is denoted as $S^t \in [0,1]^{hw \times hw}$. 
Similarly, we use $S^t_p \in [0,1]^{h \times w}$ to denote the self-attention map of all pixels attending to pixel $p$.

\vspace{-0.1cm}
\subsection{Grounded text-to-image generation}
\label{sec:method_txt2img}

We now introduce two losses on the attention layers to improve the controllability of the layout-conditioned image synthesis. 
We consider spatial layouts defined by $k$ bounding boxes $B\in (\mathbb{Z^+})^{k\times 4}$. 
Each box is associated with a box caption describing the content inside the box. 
Given the captions associated with every box, we denote their indices in the input text prompt $\mathbf{w}$ as $I = \{i_1 \cdots i_q\}$ ($q$ is the number of those tokens of interest). 
Note that each token index $i$ can relate to one or more bounding boxes $B_i$. 
Let $\text{Mask}(B_i)$ be the binary mask generated from the boxes $B_i$, where the regions inside the boxes are one, and the rest are zero, as shown in the fourth row of ~\Fref{fig:framework}.


\vspace{-0.2cm}
\subsubsection{Cross-Attention Refocusing (CAR)}
\vspace{-0.1cm}
When generating images using GLIGEN, given the text prompt ``three parrots'', we notice in the cross-attention layer that the token ``parrot`` incorrectly attends to the unrelated regions, as visualized in~\Fref{fig:motivation}a, leading to four parrots generated in the result.
To this end, we propose a loss to \emph{refocus} the cross-attention of these tokens according to the layout. 

\myparagraph{Post-processing cross-attention maps.} 
First, we skip the attention maps of $<sot>$ token, then use $Softmax$ for the remaining cross-attention maps. 
After that, we apply Gaussian Smoothing to the attention maps following~\cite{chefer2023attend}.

\myparagraph{Designing cross-attention refocusing loss.}
 To encourage the generated objects to present in corresponding boxes, we aim to boost the scores of masked attention map 
   $A_i^t \cdot \text{Mask}(B_i)$:
\begin{equation}
 \vspace{-0.1cm}
    \mathcal{L}_{FG} = \frac{1}{q}  \sum_{i\in I} (1 - \max(A_i^t \cdot \text{Mask}(B_i)))
 \vspace{-0.1cm}
\end{equation}







We also propose a background loss to discourage the tokens from being attended by the irrelevant regions:
\begin{equation}
    \mathcal{L}_{BG} = \frac{1}{q} \sum_{i \in I} \max(A_i^t \cdot (1 - \text{Mask}(B_i)))
\vspace{-0.1cm}
\end{equation}



The overall CAR loss is then defined as $\mathcal{L}_{CAR} = \mathcal{L}_{FG} + \mathcal{L}_{BG}$. 
As shown in \Fref{fig:motivation}a, when applying with our loss, the model effectively mitigates the incorrect attention to the grounded tokens and synthesizes three parrots as desired.

Adopting CAR loss for segmentation mask can be found in Appendix ~\Sref{sec:adding_seg}


\vspace{-0.5em}
\subsubsection{Self-Attention Refocusing (SAR)}

Similar to the observation in the cross-attention layers, as shown in \Fref{fig:motivation}b, the pixels of one region (e.g., ``car'') may attend \emph{outside} of the region to similar regions (e.g., ``chair'') in self-attention layers. 
As a result, the attributes of the two regions get mixed in the generation.
To this end, we develop a loss to help self-attention \emph{refocus} to the correct regions. 

Recall that $S^t_{p}$ is the self-attention map of pixel $p$. 
For each pixel $p \in B_i $, we denote $S^{t,BG}_{p}$ as the background region of the self-attention map:
\begin{equation}
    S^{t,BG}_{p} = S^t_{p} \cdot (1 - \text{Mask}(B_i))
\end{equation}

We aim to ensure each pixel $p \in B_i$ attends less to the regions outside the boxes $B_i$. 
To achieve this, we define self-attention loss for each pixel $p$ as follows: 
\begin{equation}
\mathcal{L}_{p} = \frac{\sum (S^{t,BG}_{p})}{\sum (1 - \text{Mask}(B_i))} 
 \end{equation}
The overall self-attention loss is defined as follows:
\begin{equation}
\mathcal{L}_{SAR} = \frac{1}{q}  \sum_{i \in I} \sum_{p \in B_i} \mathcal{L}_{p}
 \end{equation}


As shown in ~\Fref{fig:motivation}b, using self-attention loss helps each box to focus less on the irrelevant regions, and the model consequently generates distinct attributes for each region.

\vspace{-0.5em}
\subsubsection{Sampling with the attention-refocusing losses}

With the CAR and SAR losses, we modify the noised sample $x_t$ at each denoising step to minimize the loss using gradient descent.
We show the update process in \Fref{fig:framework}:
\vspace{-0.2cm}
\begin{equation}
\label{eq:update}
    \hat{x}_{t} \leftarrow {x}_{t} - \alpha \nabla_{{x}_{t}} \left(\mathcal{L}_{CAR} + \mathcal{L}_{SAR}\right),
 \vspace{-0.1cm}
\end{equation}
where $\alpha$ is the step size that controls the influence of the optimization in the denoising process. 
However, a single update step is often insufficient to refine the cross-attention and self-attention maps. 
We thus update $\tau$ times every early denoising step. 
After finishing the $\tau$ updates, we feed the output to the diffusion UNet to resume the denoising process and compute $x_{t-1}$.
Intuitively, we use the gradient derived from attention-refocusing losses to guide the denoising process.
More details about $\tau$ setting and algorithm can be found in the Appendix~\Sref{sec:adding_set_up}.
\vspace{-0.2em}

\subsection{Text-to-layout prediction}
\label{sec:method_layout}
 
Generating an image from text requires strong text comprehension and reasoning capacity.  
The limited power of text encoders could be another reason existing methods fail.
However, once a text-to-image model is trained with a specific language model, upgrading the text encoder without additional (costly) training becomes non-trivial. 
Such schema could hinder the existing text-to-image models from benefiting from recent large language models (LLMs) breakthroughs.
Given this challenge, we explore directly using LLMs to generate intermediate visual presentations such as box layouts.

We exploit GPT-4~\cite{openai2023gpt4}, the state-of-the-art large language model that can understand the number and spatial compositions of objects in our experiments. 
Specifically, given the input text for image generation, we create a new prompt to request GPT-4 to generate box coordinates and the label of objects in each box. 
We outline the details of this in-context learning in the Appendix ~\Sref{sec:gpt-4}
\vspace{-0.2cm}

\section{Experiments}
\label{sec:experiment}
\vspace{-0.1cm}
We evaluate our methods on several benchmarks, conduct ablation experiments on each component.

\begin{table}[t!]
    \centering
    \vspace{-0.2cm}
    \footnotesize{
    \tabulinesep=0pt
    
    \begin{tabu} to 1.\columnwidth {@{}X[1.8, l]X[0.6,c]X[1.2,c]X[1.3,c]X[1.3,c]X[1.3,c]@{}}
        \toprule
        \multirow{2}{*}{\textbf{Method}} & \multirow{2}{*}{\shortstack[c]{\hspace{-0.3cm}\textbf{CAR} \\ \hspace{-0.3cm}\& \textbf{SAR}}} &\textbf{Counting}&\multicolumn{3}{c}{ \textbf{Compositions}}\\
         \cmidrule(r){3-3} \cmidrule(r){4-6}
         \footnotesize & &\hspace{-0.2cm} F1~$\uparrow$ & \hspace{-0.2cm}Spatial~$\uparrow$ & \hspace{-0.45cm}Size~$\uparrow$& \hspace{-0.3cm}Color~$\uparrow$ \\
        \midrule
        \multirow{2}{*}{\shortstack[l]{Stable Diffusion \\ \cite{rombach2022high}}} & \xmark &\hspace{-0.4cm}58.31 &  \hspace{-0.3cm}8.48 &\hspace{-0.5cm} 9.18 & \hspace{-0.6cm}12.61 \\
        & \cmark &  60.62~\goodi{2.3} & 24.45~\goodi{16.0} & 16.97~\goodi{7.7} &23.54~\goodi{10.9}  \\
        \midrule
        \multirow{2}{*}{\shortstack[l]{Attend-and-excite \\ \cite{chefer2023attend}}} & \xmark & \hspace{-0.4cm}60.47&  \hspace{-0.3cm}9.98 & \hspace{-0.5cm}10.58 &\hspace{-0.6cm} 19.56   \\
        & \cmark & 62.71~\goodi{2.2} & 20.76~\goodi{10.8} & 14.17~\goodi{3.6} & 20.83~\goodi{1.3} \\
        \midrule
        \multirow{2}{*}{\shortstack[l]{Layout-guidance \\ \cite{chen2023training}}} & \xmark & \hspace{-0.4cm}56.22 &\hspace{-0.5cm}16.47 &\hspace{-0.5cm} 12.38 & \hspace{-0.6cm}14.39    \\
        & \cmark & 63.01~\goodi{6.8} & 25.84~\goodi{9.4} & 15.56~\goodi{3.2} & 21.50~\goodi{7.1} \\
        \midrule
        \multirow{2}{*}{\shortstack[l]{MultiDiffusion \\ \cite{bar2023multidiffusion}}} & \xmark & \hspace{-0.4cm}55.18 &\hspace{-0.5cm} 14.27 &\hspace{-0.5cm} 10.58 & \hspace{-0.6cm}17.15   \\
        &\cmark & 57.37~\goodi{2.19} & 22.65~\goodi{8.2} & 10.78~\goodi{0.2} & 24.59~\goodi{7.3}  \\
        \midrule
        \multirow{2}{*}{\shortstack[l]{GLIGEN  \cite{li2023gligen}}} & \xmark & \hspace{-0.4cm}66.58 
 &\hspace{-0.5cm}30.74 & \hspace{-0.5cm}26.75  &\hspace{-0.6cm} 18.78  \\
        & \cmark & 67.54~\goodi{0.7} & 40.22~\goodi{9.5} &  27.74~\goodi{1.0} &  26.32~\goodi{7.5} \\
        \bottomrule
    \end{tabu}
    }
    \vspace{-0.5em}
    \caption{The CAR and SAR losses increase the F1 score in counting and  accuracy(\%) in all spatial, size, and color categories of the HRS benchmark.}
    \label{tab:spatial_counting_hrs}
    \vspace{-0.3em}
\end{table}

\subsection{Experiment setup}
\label{sec:exp_detail}
\vspace{-0.2cm}

\begin{table}[t]
    \centering

    \footnotesize{
    \tabulinesep=0pt
    \begin{tabu} to 1.\columnwidth {@{}X[2, l]X[0.8,c]X[1.6,c]X[1.6,c]X[1.6,c]@{}}
        \toprule
        \textbf{Method} &\shortstack[c]{\textbf{CAR} \\ \& \textbf{SAR}} &\textbf{Counting} &\textbf{Spatial} & \textbf{Average}\\

        \midrule
        \multirow{2}{*}{\shortstack[l]{Stable Diffusion 1.4\\ \cite{rombach2022high}}} & \xmark &  \hspace{-0.5cm} 68 .15 & \hspace{-0.5cm} 72.01   &  \hspace{-0.5cm} 78.38 \\
        & \cmark & \hspace{0.1cm}69.37~\goodi{1.22}   & \hspace{0.1cm}73.33~\goodi{1.32}  &  \hspace{0.1cm} 78.87~\goodi{0.49} \\
        \midrule


        \multirow{2}{*}{\shortstack[l]{Stable Diffusion 2.1\\ \cite{rombach2022high}}} & \xmark &  \hspace{-0.5cm} 73.63   & \hspace{-0.5cm} 76.11  &  \hspace{-0.5cm} 81.84  \\
        & \cmark & \hspace{0.1cm}74.44~\goodi{0.81}   & \hspace{0.1cm}76.29~\goodi{0.18}  &  \hspace{0.1cm} 81.89~\goodi{0.05} \\
        \bottomrule
    \end{tabu}}
    \vspace{-0.7em}
    \caption{TIFA score(~$\uparrow$) in two baselines: Stable Diffusion 1.4 and 2.1.}
    \label{tab:tifa}
    \vspace{-0.3em}
\end{table}

\begin{table}[t]
    \centering
    \footnotesize{
    \tabulinesep=0pt
    \begin{tabu} to 1.\columnwidth {@{}X[3.2, l]X[1.8,c]X[2.2,c]X[1.6,c]@{}}
        \toprule
        \textbf{Method} &\textbf{IoU} $\uparrow$ &\textbf{Clip Score} $\uparrow$ & \textbf{SOA-I} $\uparrow$\\
        \midrule
       
          SD-Pww ~\cite{balaji2022ediffi} & \hspace{-0.4cm}$23.76 \pm 0.50$ & \hspace{-0.33cm}$0.2800 \pm 0.0005 $&\hspace{-0.25cm}$73.92 \pm 1.84$ \\
         DenseDiffusion~\cite{kim2023dense} & \hspace{-0.4cm}34.99 $\pm$ 1.13& \hspace{-0.33cm}0.2814 $\pm$ 0.0005 &\hspace{-0.25cm}77.61 $\pm$ 1.75 \\
         Stable Diffusion + Our &\hspace{-0.4cm}\textbf{38.97 $\pm$ 0.56} & \hspace{-0.33cm}\textbf{0.3177 $\pm$ 0.0011}&\hspace{-0.25cm}\textbf{78.80 $\pm$ 1.27}\\
    
        \bottomrule
    \end{tabu}}
    \vspace{-0.7em}
    \caption{Evaluation of image generation based on mask guidance, highlighting the performance of our approach. The results of other methods are directly taken from DenseDiffusion ~\cite{kim2023dense}.}
     \label{tab:mask}
    \vspace{-0.3em}
\end{table}

\begin{table}[t]
    \centering
    \label{tab:fid}
    \footnotesize{
    \begin{tabu} to 1.\columnwidth {@{}X[1.7, l]X[0.7,c]X[1.3,c]X[1.1,c]X[1.3,c]X@{}}
        \toprule
     & \multicolumn{2}{c} {\textbf{Stable Diffusion} ~\cite{rombach2022high}} & \multicolumn{2}{c}{\textbf{GLIGEN} ~\cite{li2023gligen}}\\
      \cmidrule(r){2-3} \cmidrule(r){4-5}
      \textbf{CAR\&SAR} & \xmark & \cmark & \xmark & \cmark\\
      \midrule
       &20.82 & 21.03 & 20.63 & 20.37 \\ 
      \bottomrule
    \end{tabu}}
    \vspace{-0.7em}
    \caption{FID ( $\downarrow$) in Stable Diffusion and GLIGEN with and without CAR \& SAR in COCO 2014~\cite{cocodataset}}
    \vspace{-1em}
\end{table}
\myparagraph{Dataset.} 
For text-to-image tasks, we utilize the benchmark \tb{HRS} ~\cite{bakr2023hrs} and \tb{Drawbench} ~\cite{Imagen} to evaluate the text-to-image generation performance on various categories, including counting, spatial, color, and size compositions. 
To further assess the alignment of the generated image and input text, we use \tb{TIFA} ~\cite{hu2023tifa} benchmark. 
\tb{The HRS dataset} contains various prompts divided into three main categories: 1) accuracy, 2) robustness, and 3) generalization. 
Our method focuses on \emph{accuracy improvement}, including four main categories: \emph{spatial relationship, color, size, and counting}.
Each prompt in the dataset is tagged with the object's name and corresponding labels intended for evaluation. 
For example, in spatial relationships, the labels include objects and their relative positions, such as ``on the left'' or ``on the right''. 
The prompts for each category counting/spatial/size/color are $3,000/1,002/501/501$. 
Depending on the number of objects and their relationship, we label the difficulty level of each prompt as easy, medium, and hard with roughly the same amount. 
\tb{The DrawBench dataset} consists of 39 prompts about \emph{Counting} and \emph{Positional (or spatial relationship)}. 
Since there are no labels for this benchmark, we manually create the label for each prompt based on the number of objects mentioned and their relationships. 
\tb{The TIFA benchmark} contains $4,000$ prompts in various categories (counting, spatial, food, locations, etc...) and the questions for each prompt, along with their answers.

For quality evaluation, we utilize the \tb{COCO2014} ~\cite{cocodataset} validation dataset to assess images generated from textual descriptions and corresponding bounding boxes.

Our mask-and-text-to-image evaluation utilizes the dataset provided by DenseDiffusion~\cite{kim2023dense}, which includes about 250 binary masks with corresponding labels and captions, allowing us to evaluate our method's capability in adhering to the provided mask guidances. 

\begin{figure}[t]
\vspace{-0.2cm}
\centering
    \begin{subfigure}[t]{0.02\textwidth}
        \vspace{-4.9cm}
       
        {
        \begin{tabu} to 0.025\textwidth {X[c]X[c]}
        
        \rotatebox[origin=c]{90}{\hspace{0.35cm} \footnotesize{Layout}\hspace{0.7cm}}\\
        \rotatebox[origin=c]{90}{\hspace{0.cm} \footnotesize{Without ours}\hspace{0.cm}}\\
        \rotatebox[origin=c]{90}{\hspace{0.1cm} \footnotesize{With ours}\hspace{0.2cm}}\\
       \end{tabu}}
       
    \end{subfigure}
    \begin{subfigure}[t]{0.45\textwidth}
        \centering
        \includegraphics[width=1\textwidth]{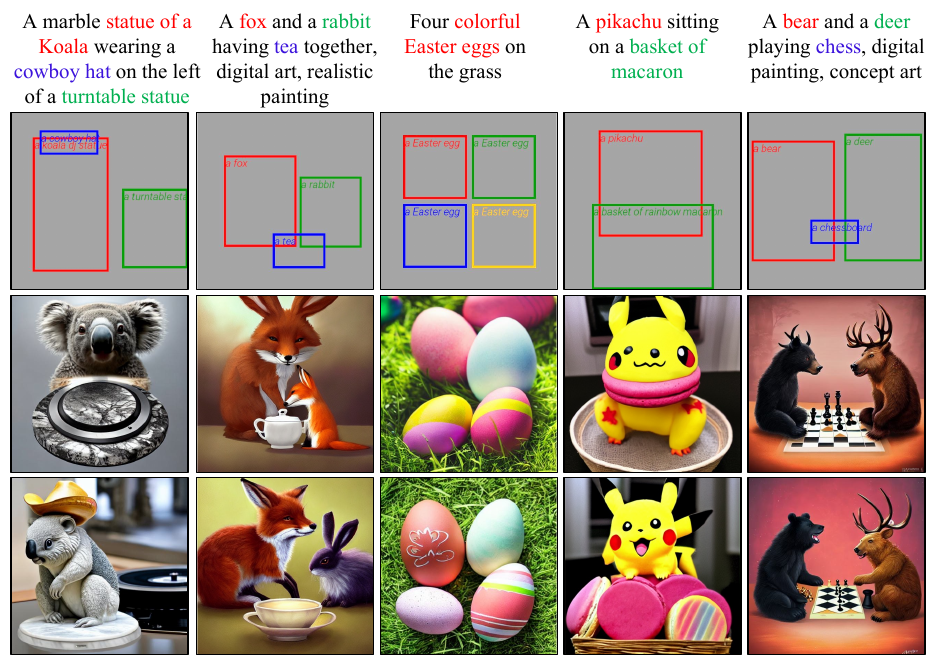}
        \vskip -0.cm
        \vspace{-0.1cm}
        {
        \begin{tabu} to 1\textwidth {X[1.2,c]X[1.2,c]X[1,c]X[1.1,c] X[1,c]}
        \hspace{-0.2cm}\footnotesize{~\cite{rombach2022high}} & \hspace{-0.1cm}\footnotesize{~\cite{chefer2023attend}} & \hspace{-0.3cm}\footnotesize{~\cite{bar2023multidiffusion}}  & \hspace{-0.2cm}\footnotesize{~\cite{chen2023training}} & \footnotesize{~\cite{li2023gligen}} \\
        \end{tabu}}
    \end{subfigure}
    \vspace{-0.3cm}
\caption{
\tb{Plug \& play use of our attention-based guidance.}
Our method applies to various base models. 
Here we show improved controllability across multiple text-to-image methods: Stable Diffusion ~\cite{rombach2022high}, Attend-and-excite~\cite{chefer2023attend}, MultiDiffusion~\cite{bar2023multidiffusion}, Layout-guidance~\cite{chen2023training}, GLIGEN ~\cite{li2023gligen}
}
\vspace{-0.5cm} 
\label{fig:update}
\end{figure}

\myparagraph{Evaluation metrics.}
Regarding text-to-image evaluation, we follow the protocol in HRS~\cite{bakr2023hrs} to compute the metrics on individual categories. 
In counting, the number of objects detected in generated images is compared to the ground truths in text prompts to calculate the precision, recall, and F1 score. 
We use accuracy as the evaluation metric for spatial, size, and color categories. False positive samples happen when the number of generated objects is smaller than the ground truths. In contrast, the false negative objects are counted for the missing objects in the synthesized images. 
For other categories, we use accuracy as the evaluation metric. 
Depending on the category, the image is counted as a correct prediction when all detected objects are correct, either for spatial relationships, color, or size.
For TIFA benchmark, we assess the alignment between the generated images and input texts using the TIFA score ~\cite{hu2023tifa}. 



In box-and-text evaluations, we report FID~\cite{heusel2017gans} in COCO2014 ~\cite{cocodataset}, using available bounding boxes from this benchmark as input layouts for grounded text-to-image models. 
We randomly choose 5k captions and corresponding images from this benchmark to calculate the FID.
We use this evaluation to validate that our generation results remain natural compared to the base model.

For mask guidance evaluation, we measure the alignment with IoU used in DenseDiffusion~\cite{kim2023dense}, assess the textual similarity with CLIP Score ~\cite{hessel2021clipscore}, and ensure the object presence using the SOA-I score ~\cite{hinz2019semantic} using YOLOv7 ~\cite{wang2023yolov7} for object detection.

\myparagraph{Implementation details.}
In terms of bounding boxes guidance, we evaluate our method by plugging it into various open-source text-to-image models and methods, including Stable Diffusion (SD) V-1.4~\cite{rombach2022high}, Attend-and-excite~\cite{chefer2023attend}, Layout-guidance~\cite{chen2023training}, MultiDiffusion~\cite{kumari2022multi}, and GLIGEN~\cite{li2023gligen}. 
For mask guidance, we integrate our losses to Stable diffusion ~\cite{rombach2022high} and compare with other free-training methods: DenseDiffusion ~\cite{kim2023dense} and Pww ~\cite{balaji2022ediffi}. 
All the mentioned methods are configured as default settings. 
More implementation details are in the Appendix ~\Sref{sec:adding_set_up}
\subsection{Quantitative results}

We show the results on the HRS benchmark ~\cite{bakr2023hrs}  in \Tref{tab:spatial_counting_hrs}. 
Using our losses consistently enhances F1 scores in counting by an average of 2\%, with Layout-guidance ~\cite{chen2023training} models showing a notable 7\% improvement. 
Spatial accuracy gains an average of 10\% with our losses. 
Our method boosts accuracy by up to 10.9\% in the size and color categories. 
Specifically, GLIGEN ~\cite{rombach2022high} sees an increase in spatial and color accuracy by around 10\% and 8\%, respectively. 
Stable Diffusion ~\cite{rombach2022high} , which relies solely on text input, lags behind other grounded text-to-image models. 
However, with our attention-guided enhancements, it outperforms several methods like Layout-guidance and MultiDiffusion. 
The TIFA evaluation in ~\Tref{tab:tifa} further demonstrates the efficacy of our CAR and SAR losses, particularly enhancing counting and spatial accuracy across all baseline versions without detriment to other categories. 
This is reflected in the overall TIFA score improvements.

Moreover, our attention-refocusing losses improve textual alignment in models like Stable Diffusion and GLIGEN without affecting image quality, maintaining FID scores compatible with the originals. 

 

For mask guidance evaluation, we report the quantitative evaluation in ~\Tref{tab:mask}. 
Our method outperforms the current state-of-the-art approaches with an IoU of 38.97 $\pm$ 0.56 and a leading SOA-I score and Clip score, indicating improved layout fidelity and object detection in generated images.

Additionally, we show the results in DrawBench ~\cite{Imagen} and compare our method with BoxDiff ~\cite{xie2023boxdiff} as well as the inference time of several free-training methods in the ~\Sref{sec:additional_quant} (Appendix)
\vspace{-0.1cm}
\begin{table}[t!]
    \centering
    \vspace{-0.2cm}
    \small{
    \tabulinesep=0pt
    \begin{tabu} to 1.\columnwidth {@{}X[0.55,c]X[0.55,c]X[1.3,c]X[1.1,c]X[0.6,c]X[1,c]X[1,c]X[1,c]@{}}
        \toprule
        \multirow{2}{*}{\textbf{CAR}} & \multirow{2}{*}{\textbf{SAR}} & \multicolumn{3}{c}{\textbf{Counting}} & \textbf{Spatial} & \textbf{Size} & \textbf{Color} \\
        \cmidrule{3-5} 
        & & Precision~$\uparrow$ & Recall~$\uparrow$ & F1~$\uparrow$ & Acc.~$\uparrow$ & Acc.~$\uparrow$ & Acc.~$\uparrow$ \\
        \midrule
        \xmark & \xmark & 78.81 & 59.44 & 66.58 & 30.74 &26.75 &18.78  \\
        \xmark& \cmark & 79.76 & 59.34 & 67.03  & 36.43 & \textbf{30.34} & 18.39  \\
        \cmark& \xmark &  \textbf{82.11} & \textbf{59.35} & \textbf{67.59} & \textbf{36.92} & \textbf{28.94} & \textbf{23.88}  \\
        \cmark& \cmark&  \textbf{81.25} & \textbf{59.39} & \textbf{67.54} & \textbf{40.22} &  27.74 & \textbf{26.32}  \\
        \bottomrule
    
    \end{tabu}}
    \vspace{-0.5em}
    \caption{\textbf{Abaltion study} of the CAR and SAR losses using the GLIGEN model on the HRS benchmark.}
    \label{tab:ablation}
   \vspace{-1em}
\end{table}

\begin{figure}[t]
\vspace{-0em}
    \centering
    \includegraphics[width=0.5\textwidth]{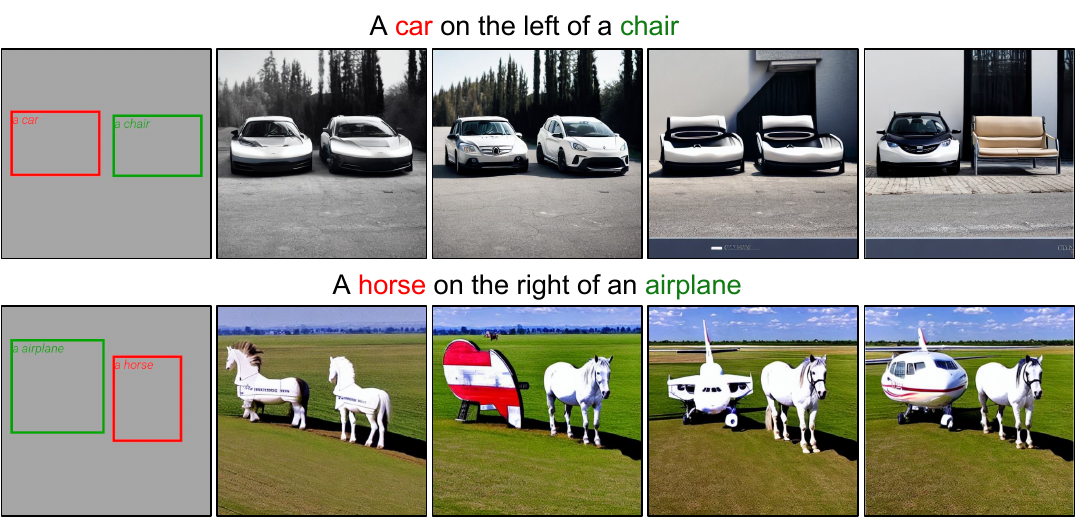}\hfill
    \begin{tabu} to 0.5\textwidth {X[1.2,l]X[1.1,c]X[1.1,c]X[1.1,c]X[1.1,c]}
        \toprule
        \footnotesize{CAR} & \footnotesize{\xmark} & \footnotesize{\xmark} & \footnotesize{\cmark} & \footnotesize{\cmark} \\
        \footnotesize{SAR} & \footnotesize{\xmark} &  \footnotesize{\cmark} & \footnotesize{\xmark} & \footnotesize{\cmark} \\
        \bottomrule
    \end{tabu}
    \vspace{-0.5em}
    \caption{
    \tb{Ablation study.} 
    We show sample grounded text-to-image generation demonstrating the effects of the two proposed attention guidance. 
    }
    \label{fig:ablation}
    \vspace{-1.2em}
\end{figure}

\label{sec:exp_quan}
\subsection{Qualitative results}

\Fref{fig:update} illustrates the qualitative comparison of various methods with and without our losses. 
Note that we generate each pair of images with the same initial noise.
In all the cases, our losses help generate images with more precise spatial locations, colors, and numbers of objects. 
For example, attention-refocusing loss helps to mitigate the attribute mixing problem of Layout-guidance ~\cite{chen2023training} (in the fourth column).

\begin{figure}[t]
    \centering
    \includegraphics[width=0.5\textwidth]{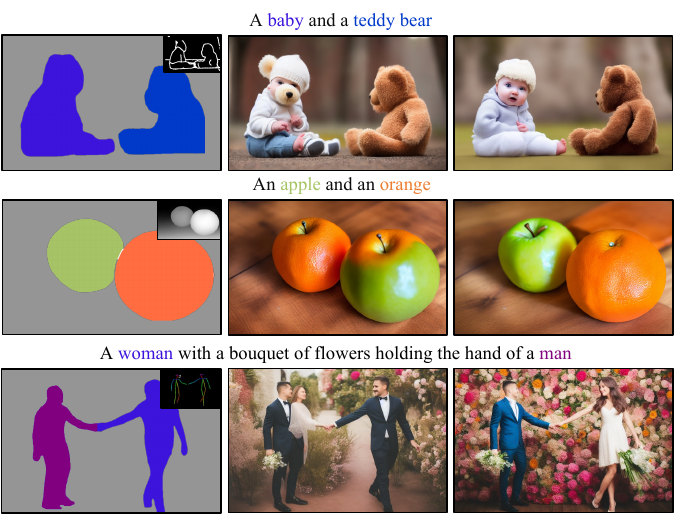}
        \begin{tabu} to 0.5\textwidth {X[2cm,c]X[1.9cm,c]X[1.9cm,c]}
        \small{Input} & \small{ControlNet ~\cite{zhang2023adding}}& \small{ControlNet + Ours} \\
        \end{tabu}
\vspace{-1.2em}
\caption{
\tb{ControlNet with attention-based guidance.}
The input of ControlNet is a small image in the top right of the first column. With an extra segmentation map (the bigger images in the first column), our losses can refine the attribute blending of ControlNet.}
\label{fig:controlnet}
\end{figure}

\begin{figure}[t]
\vspace{-0.2cm}
    \centering
    \includegraphics[width=0.5\textwidth]{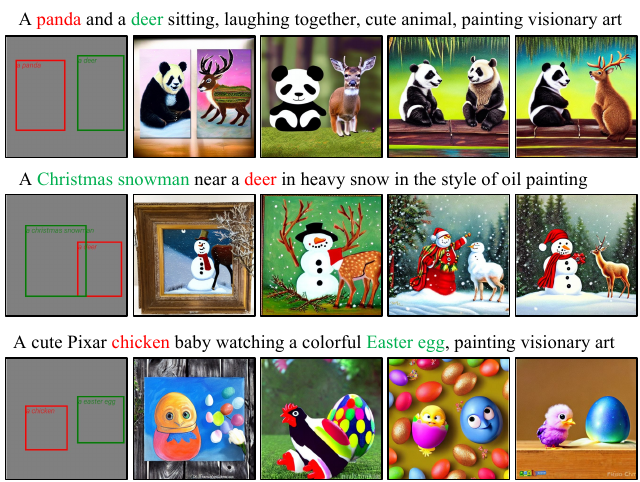}
    \begin{tabu} to 0.5\textwidth {X[0.8,c]X[1.0,c]X[0.9,c]X[0.8,c]X[0.8,c]}
        \rule{0pt}{-1cm} \footnotesize{Layout from GPT} & \footnotesize{Layout-guidance~\cite{chen2023training}}  &
        \hspace{-0.1cm}\footnotesize{MultiDiffusion} ~\cite{kumari2022multi}   & \footnotesize{GLIGEN} ~\cite{li2023gligen} & \footnotesize{GLIGEN + Ours}
    \end{tabu}
    \vspace{-0.3cm}
    \caption{
    \tb{Visual comparisons on HRS benchmark.}
    Here, we apply our attention-based guidance to grouned text-to-image models.
    All methods take the same grounded texts as inputs.
    The results show the capability of our method in synthesizing novel spatial compositions and attributes.
    }
\label{fig:4baseline}
\vspace{-1.5em}
\end{figure}

In ~\Fref{fig:4baseline},  we show the results using prompts from the HRS benchmark. 
While MultiDiffusion ~\cite{kumari2022multi} and Layout-guidance ~\cite{chen2023training} often do not respect the input layouts, particularly in smaller boxes, GLIGEN incorrectly aligns objects with grounded input or mismatch colors. 
In contrast, integrating our attention-refocusing losses with GLIGEN enhances alignment and color accuracy. 
The results of our approach underscore our method's effectiveness in creating novel spatial configurations and attributes. More results can be found in Appendix ~\Fref{fig:additional_baselines} (comparing with four baselines), and ~\Sref{sec:additional_plug} (comparing with GLIGEN)


Additionally, as shown in ~\Fref{fig:controlnet}, our CAR and SAR losses can adapt to segmentation mask guidance, refining ControlNet's output by reducing attribute mixing and avoiding generating additional and irrelevant objects.
\vspace{-0.1cm}
\subsection{Ablation studies}
\label{sec:exp_abl}
\vspace{-0.1cm}
We ablate the two losses using GLIGEN~\cite{li2023gligen} as the baseline method in ~\Fref{fig:ablation}. 
GLIGEN sometimes struggles with prompts with multiple objects, especially objects in the same category or size. 
CAR loss can mitigate this problem, but the generated objects still have attributes blended from others. 
For instance, with only CAR loss, a generated car might have a mixed chair feature and vice versa. 
Incorporating CAR and SAR losses further mitigates the attribute blending problem.

We further perform quantitative evaluation in ~\Tref{tab:ablation} for all categories in the HRS benchmark. 
Adding CAR or SAR loss to the GLIGEN model improves the baseline in all four categories. 
Particularly in spatial relationships, using SAR or SAR can improve GLIGEN by approximately 6\%.
When using both losses, we can achieve an around 10\% accuracy improvement.

\vspace{-0.1cm}
\subsection{Large language model evaluation}
\label{sec:llm}

We examine the latest large language models, GPT-4 ~\cite{openai2023gpt4} GPT-3 ~\cite{brown2020language}, Llama 1 ~\cite{touvron2302llama} and Llama 2 ~\cite{touvron2307llama} (version 13b-chat) by evaluating their ability to comprehend the visual concept. 
We randomly chose 200 prompts from four categories in the HRS benchmark and report three metrics: 
\begin{itemize}

\item Format: whether or not the model returns the correct format of grounded information, including four coordinates for each box along with its label.

\item Validness: all generated boxes are satisfied with the size and box constraints, eg. the coordinate box is $\{x_1,y_1,x_2,y_2\}$ then $512 \ge x_1, x_2, y_1, y_2 \geq 0$,  $x_1 \leq x_2 $ and $y_1 \leq y_2$,

\item Correctness: the generated grounding information should follow the text prompts. For example, in terms of counting, the quantity of generated boxes should match the number of objects mentioned in the input prompt. In spatial and size categories, we asses the relations and relative size of generated boxes. Meanwhile, in color, we verify if the correct colors are returned for each object in the grounding text.
\end{itemize}

\begin{table}[t]
\vspace{-0.2cm}
\centering
\begin{tabu} to 0.45\textwidth {X[1.7,l]X[1.1,c]X[1.2,c]X[1.1,c]}
\toprule
{Model} & {Format} $\uparrow$ & {Valid} $\uparrow$ & {Correct} $\uparrow$\\ 
\midrule
Llama 1 ~\cite{touvron2302llama} & 67.5 & 46.0 & 38.5 \\
Llama 2 ~\cite{touvron2307llama} & 98.5 & 84.0 & 63.5 \\ 
GPT-3 ~\cite{brown2020language} & \textbf{98.5} & 97.5 & 83.5 \\ 
GPT-4 ~\cite{openai2023gpt4} & \textbf{98.5} & \textbf{98.5} & \textbf{88.5} \\ 
\bottomrule
\end{tabu}
\vspace{-0.2cm}
\caption{Performance evaluation of LLMs using 200 random prompts in the HRS benchmark ($\%$) }
\label{tab:llm}
\vspace{-0.3cm}
\end{table}

\begin{figure}[t]
    \centering
    \includegraphics[width=0.5\textwidth]{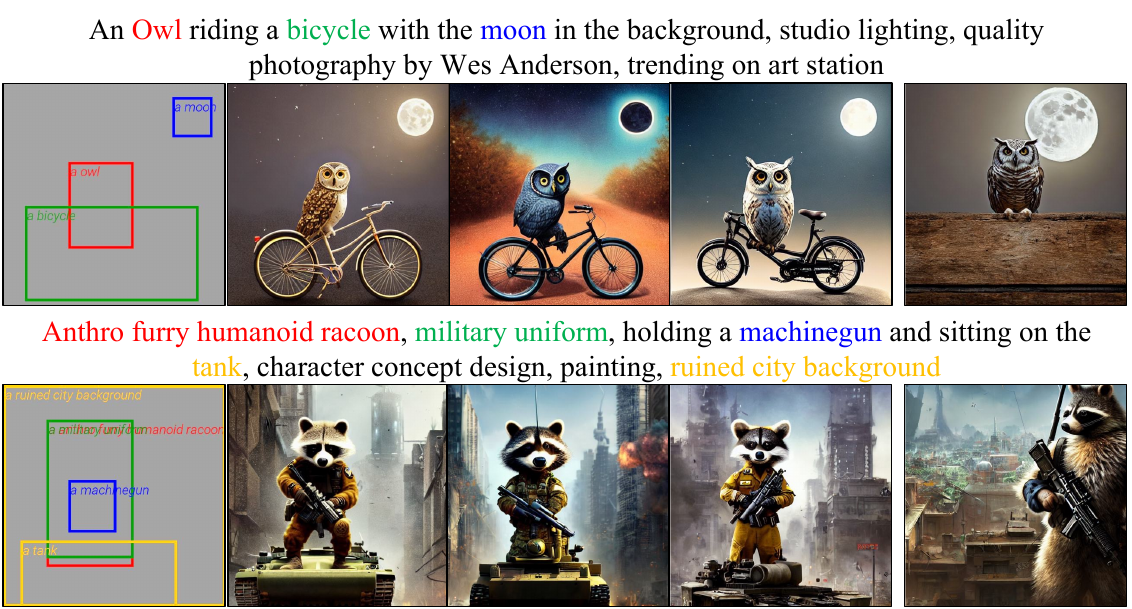}
    \begin{tabu} to 0.5\textwidth {X[0.9,c]X[3.1,c]X[1.2,c]}
        \rule{0pt}{-1cm} \footnotesize{Layout from GPT} & \footnotesize{GLIGEN + Ours} &\hspace{0.1cm} \footnotesize{Stable Diffusion}~\cite{rombach2022high}
    \end{tabu}
    \vspace{-0.4cm}
    \caption{\textbf{Comparisons of Stable Diffusion and our two-stage pipeline.}
    Our two-stage pipeline excels over Stable Diffusion ~\cite{rombach2022high} in prompt understanding. }
    
    \label{fig:gpt4_visual}
\vspace{-1em}
\end{figure}

In ~\Tref{tab:llm}, GPT-4 outperforms Llama 1, Llama 2, and GPT-3 in three metrics. The visual comparisons are shown in ~\Fref{fig:llama_gpt} (Appendix).
Leveraging the leading large language model, our two-stage text-to-image model surpasses the single-stage Stable Diffusion ~\cite{rombach2022high} in understanding object relationships and textual alignment, as shown in ~\Fref{fig:gpt4_visual}.

        
    
    


\subsection{Instructing text-to-image by chatGPT}
\label{sec:instruct}

\begin{figure}[t]
    \centering
    \includegraphics[width=0.5\textwidth]{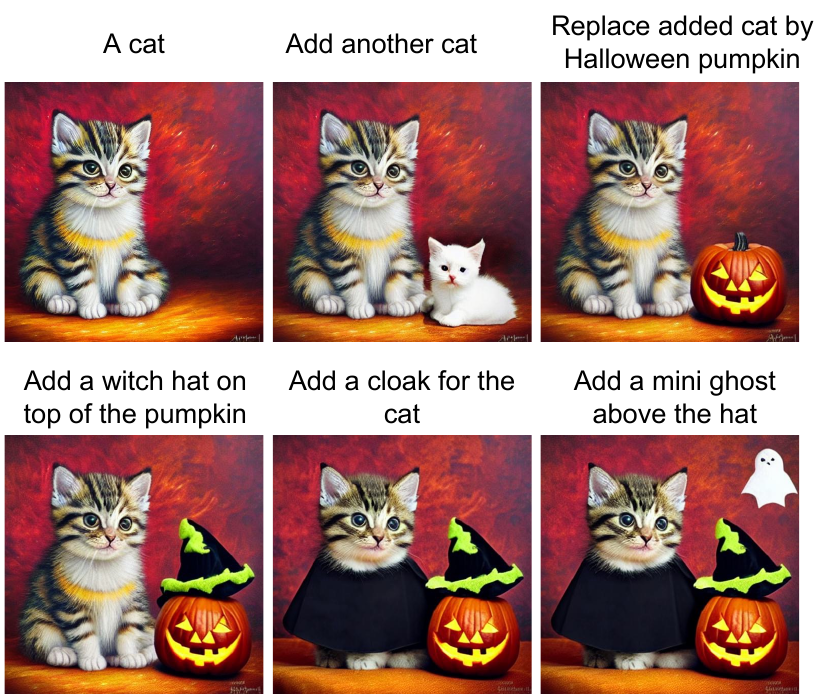}
    \caption{ Instruct text-to-image by instructing chatGPT.}
    \label{fig:instruct}
    \vspace{-0.4cm}
\end{figure}

We also propose a novel capability enabled by our framework, where users can utilize chatGPT to instruct text-to-image. 
In other words, after generating the initial layout and image, we instruct chatGPT to modify the layout, leading to an updated image. 
This iterative capability allows users to synthesize desired images through consecutive adjustments. 
As shown in ~\Fref{fig:instruct}, a user wants to generate a Halloween-themed image. Initially, the user generates a cat and adds another cat to its right. Unsatisfied with this, they replace the second cat with a Halloween pumpkin. They continue to add a witch hat on the pumpkin, a cloak for the cat, and a mini ghost in the background for a playful touch.
Such language-based refinement ability is difficult for traditional text-to-image models to offer.



\section{Limitation}
\begin{figure}[t]
    \centering
    \includegraphics[width=0.5\textwidth]{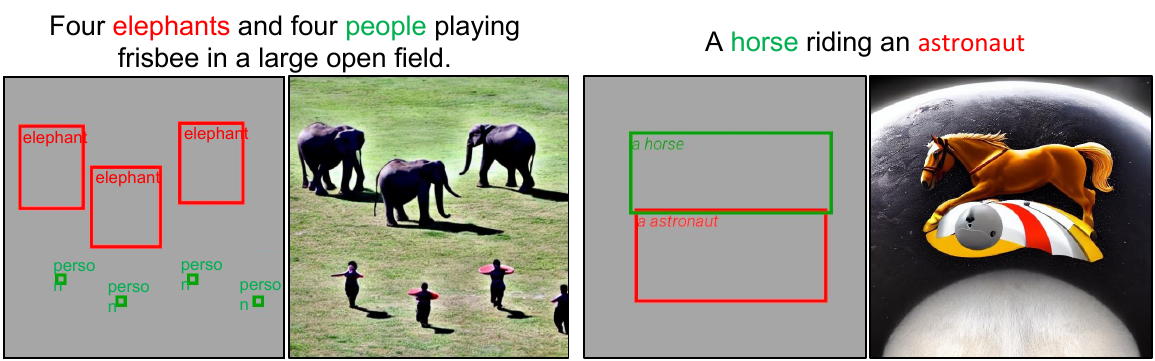}
    \vspace{-1.4em}
    \caption{\textbf{The failures cases of our framework. }
    GPT-4 sometimes misinterprets object quantity or size and instances of the text-to-image model not aligning with GPT-4's layout}
    
    \label{fig:limitation}
\vspace{-0.5cm}
\end{figure}

\vspace{-0.1cm}
\label{sec:limit}
The ~\Fref{fig:limitation} illustrates failure cases where our framework struggles. When dealing with prompts describing a large number of objects, GPT-4 occasionally produces an incorrect count or generates small boxes. Additionally, there are instances where GPT-4 accurately generates the layout, yet the grounded text-to-image model fails to adhere to these out-of-distribution layouts (the second example).
\vspace{-0.2cm}

\section{Conclusion}
In this paper, we propose a novel attention-refocusing approach to improve the alignment of cross- and self-attention layers given layouts during the sampling process. 
Furthermore, we explore the usage of Large Language Models for generating visual layouts from text prompts.
Our proposed losses can be easily incorporated into existing text-to-image diffusion models.
The comprehensive experiments show favorable performance against state-of-the-art grounded text-to-image models.



\label{sec:conclucion}

\section{Acknowledgments}

We thank Hadi Alzayer and Chuong Huynh for their helpful discussion and paper reading. 
{
    \small
    \bibliographystyle{ieeenat_fullname}
    \bibliography{main}

\begin{thebibliography}{61}
\providecommand{\natexlab}[1]{#1}
\providecommand{\url}[1]{\texttt{#1}}
\expandafter\ifx\csname urlstyle\endcsname\relax
  \providecommand{\doi}[1]{doi: #1}\else
  \providecommand{\doi}{doi: \begingroup \urlstyle{rm}\Url}\fi

\bibitem[Avrahami et~al.(2023)Avrahami, Hayes, Gafni, Gupta, Taigman, Parikh,
  Lischinski, Fried, and Yin]{avrahami2022spatext}
Omri Avrahami, Thomas Hayes, Oran Gafni, Sonal Gupta, Yaniv Taigman, Devi
  Parikh, Dani Lischinski, Ohad Fried, and Xi Yin.
\newblock Spatext: Spatio-textual representation for controllable image
  generation.
\newblock \emph{CVPR}, 2023.

\bibitem[Bakr et~al.(2023)Bakr, Sun, Shen, Khan, Li, and
  Elhoseiny]{bakr2023hrs}
Eslam~Mohamed Bakr, Pengzhan Sun, Xiaogian Shen, Faizan~Farooq Khan, Li~Erran
  Li, and Mohamed Elhoseiny.
\newblock Hrs-bench: Holistic, reliable and scalable benchmark for
  text-to-image models.
\newblock In \emph{Proceedings of the IEEE/CVF International Conference on
  Computer Vision}, pages 20041--20053, 2023.

\bibitem[Balaji et~al.(2022)Balaji, Nah, Huang, Vahdat, Song, Kreis, Aittala,
  Aila, Laine, Catanzaro, et~al.]{balaji2022ediffi}
Yogesh Balaji, Seungjun Nah, Xun Huang, Arash Vahdat, Jiaming Song, Karsten
  Kreis, Miika Aittala, Timo Aila, Samuli Laine, Bryan Catanzaro, et~al.
\newblock ediffi: Text-to-image diffusion models with an ensemble of expert
  denoisers.
\newblock \emph{arXiv preprint arXiv:2211.01324}, 2022.

\bibitem[Bansal et~al.(2023)Bansal, Chu, Schwarzschild, Sengupta, Goldblum,
  Geiping, and Goldstein]{bansal2023universal}
Arpit Bansal, Hong-Min Chu, Avi Schwarzschild, Soumyadip Sengupta, Micah
  Goldblum, Jonas Geiping, and Tom Goldstein.
\newblock Universal guidance for diffusion models.
\newblock \emph{arXiv preprint arXiv:2302.07121}, 2023.

\bibitem[Bar-Tal et~al.(2023)Bar-Tal, Yariv, Lipman, and
  Dekel]{bar2023multidiffusion}
Omer Bar-Tal, Lior Yariv, Yaron Lipman, and Tali Dekel.
\newblock Multidiffusion: Fusing diffusion paths for controlled image
  generation.
\newblock 2023.

\bibitem[Brown et~al.(2020)Brown, Mann, Ryder, Subbiah, Kaplan, Dhariwal,
  Neelakantan, Shyam, Sastry, Askell, et~al.]{brown2020language}
Tom Brown, Benjamin Mann, Nick Ryder, Melanie Subbiah, Jared~D Kaplan, Prafulla
  Dhariwal, Arvind Neelakantan, Pranav Shyam, Girish Sastry, Amanda Askell,
  et~al.
\newblock Language models are few-shot learners.
\newblock \emph{Advances in neural information processing systems},
  33:\penalty0 1877--1901, 2020.

\bibitem[Byeon et~al.(2022)Byeon, Park, Kim, Lee, Baek, and
  Kim]{kakaobrain2022coyo-700m}
Minwoo Byeon, Beomhee Park, Haecheon Kim, Sungjun Lee, Woonhyuk Baek, and
  Saehoon Kim.
\newblock Coyo-700m: Image-text pair dataset.
\newblock \url{https://github.com/kakaobrain/coyo-dataset}, 2022.

\bibitem[Chang et~al.(2023)Chang, Zhang, Barber, Maschinot, Lezama, Jiang,
  Yang, Murphy, Freeman, Rubinstein, Yuanzhen, and Dilip]{chang2023muse}
Huiwen Chang, Han Zhang, Jarred Barber, AJ Maschinot, Jose Lezama, Lu Jiang,
  Ming-Hsuan Yang, Kevin Murphy, William~T Freeman, Michael Rubinstein, Li
  Yuanzhen, and Krishnan Dilip.
\newblock Muse: Text-to-image generation via masked generative transformers.
\newblock \emph{arXiv preprint arXiv:2301.00704}, 2023.

\bibitem[Chefer et~al.(2023)Chefer, Alaluf, Vinker, Wolf, and
  Cohen-Or]{chefer2023attend}
Hila Chefer, Yuval Alaluf, Yael Vinker, Lior Wolf, and Daniel Cohen-Or.
\newblock Attend-and-excite: Attention-based semantic guidance for
  text-to-image diffusion models.
\newblock 2023.

\bibitem[Chen et~al.(2023)Chen, Laina, and Vedaldi]{chen2023training}
Minghao Chen, Iro Laina, and Andrea Vedaldi.
\newblock Training-free layout control with cross-attention guidance.
\newblock \emph{arXiv preprint arXiv:2304.03373}, 2023.

\bibitem[Cho et~al.(2023)Cho, Zala, and Bansal]{cho2023visual}
Jaemin Cho, Abhay Zala, and Mohit Bansal.
\newblock Visual programming for text-to-image generation and evaluation.
\newblock \emph{arXiv preprint arXiv:2305.15328}, 2023.

\bibitem[Ding et~al.(2022)Ding, Zheng, Hong, and Tang]{ding2022cogview2}
Ming Ding, Wendi Zheng, Wenyi Hong, and Jie Tang.
\newblock Cogview2: Faster and better text-to-image generation via hierarchical
  transformers.
\newblock \emph{arXiv preprint arXiv:2204.14217}, 2022.

\bibitem[Feng et~al.(2023{\natexlab{a}})Feng, He, Fu, Jampani, Akula, Narayana,
  Basu, Wang, and Wang]{feng2022training}
Weixi Feng, Xuehai He, Tsu-Jui Fu, Varun Jampani, Arjun Akula, Pradyumna
  Narayana, Sugato Basu, Xin~Eric Wang, and William~Yang Wang.
\newblock Training-free structured diffusion guidance for compositional
  text-to-image synthesis.
\newblock In \emph{ICLR}, 2023{\natexlab{a}}.

\bibitem[Feng et~al.(2023{\natexlab{b}})Feng, Zhu, Fu, Jampani, Akula, He,
  Basu, Wang, and Wang]{feng2023layoutgpt}
Weixi Feng, Wanrong Zhu, Tsu-jui Fu, Varun Jampani, Arjun Akula, Xuehai He,
  Sugato Basu, Xin~Eric Wang, and William~Yang Wang.
\newblock Layoutgpt: Compositional visual planning and generation with large
  language models.
\newblock In \emph{ICLR}, 2023{\natexlab{b}}.

\bibitem[Gafni et~al.(2022)Gafni, Polyak, Ashual, Sheynin, Parikh, and
  Taigman]{Make-A-Scene}
Oran Gafni, Adam Polyak, Oron Ashual, Shelly Sheynin, Devi Parikh, and Yaniv
  Taigman.
\newblock Make-a-scene: Scene-based text-to-image generation with human priors.
\newblock In \emph{ECCV}, pages 89--106. Springer, 2022.

\bibitem[Ge et~al.(2023)Ge, Park, Zhu, and Huang]{ge2023expressive}
Songwei Ge, Taesung Park, Jun-Yan Zhu, and Jia-Bin Huang.
\newblock Expressive text-to-image generation with rich text.
\newblock \emph{arXiv preprint arXiv:2304.06720}, 2023.

\bibitem[Hessel et~al.(2021)Hessel, Holtzman, Forbes, Bras, and
  Choi]{hessel2021clipscore}
Jack Hessel, Ari Holtzman, Maxwell Forbes, Ronan~Le Bras, and Yejin Choi.
\newblock {CLIPScore:} a reference-free evaluation metric for image captioning.
\newblock In \emph{EMNLP}, 2021.

\bibitem[Heusel et~al.(2017)Heusel, Ramsauer, Unterthiner, Nessler, and
  Hochreiter]{heusel2017gans}
Martin Heusel, Hubert Ramsauer, Thomas Unterthiner, Bernhard Nessler, and Sepp
  Hochreiter.
\newblock Gans trained by a two time-scale update rule converge to a local nash
  equilibrium.
\newblock \emph{Advances in neural information processing systems}, 30, 2017.

\bibitem[Hinz et~al.(2019)Hinz, Heinrich, and Wermter]{hinz2019semantic}
Tobias Hinz, Stefan Heinrich, and Stefan Wermter.
\newblock Semantic object accuracy for generative text-to-image synthesis.
\newblock \emph{arXiv preprint arXiv:1910.13321}, 2019.

\bibitem[Ho and Salimans(2021)]{Classifier}
Jonathan Ho and Tim Salimans.
\newblock Classifier-free diffusion guidance.
\newblock In \emph{NeurIPS 2021 Workshop on Deep Generative Models and
  Downstream Applications}, 2021.

\bibitem[Ho et~al.(2020)Ho, Jain, and Abbeel]{ho2020denoising}
Jonathan Ho, Ajay Jain, and Pieter Abbeel.
\newblock Denoising diffusion probabilistic models.
\newblock \emph{Advances in Neural Information Processing Systems},
  33:\penalty0 6840--6851, 2020.

\bibitem[Ho et~al.(2022)Ho, Saharia, Chan, Fleet, Norouzi, and
  Salimans]{ho2022cascaded}
Jonathan Ho, Chitwan Saharia, William Chan, David~J Fleet, Mohammad Norouzi,
  and Tim Salimans.
\newblock Cascaded diffusion models for high fidelity image generation.
\newblock \emph{Journal of Machine Learning Research}, 23\penalty0
  (47):\penalty0 1--33, 2022.

\bibitem[Hu et~al.(2023)Hu, Liu, Kasai, Wang, Ostendorf, Krishna, and
  Smith]{hu2023tifa}
Yushi Hu, Benlin Liu, Jungo Kasai, Yizhong Wang, Mari Ostendorf, Ranjay
  Krishna, and Noah~A Smith.
\newblock Tifa: Accurate and interpretable text-to-image faithfulness
  evaluation with question answering.
\newblock \emph{arXiv preprint arXiv:2303.11897}, 2023.

\bibitem[Jim{\'e}nez(2023)]{jimenez2023mixture}
{\'A}lvaro~Barbero Jim{\'e}nez.
\newblock Mixture of diffusers for scene composition and high resolution image
  generation.
\newblock \emph{arXiv preprint arXiv:2302.02412}, 2023.

\bibitem[Kang et~al.(2023)Kang, Zhu, Zhang, Park, Shechtman, Paris, and
  Park]{kang2023gigagan}
Minguk Kang, Jun-Yan Zhu, Richard Zhang, Jaesik Park, Eli Shechtman, Sylvain
  Paris, and Taesung Park.
\newblock Scaling up gans for text-to-image synthesis.
\newblock In \emph{CVPR}, 2023.

\bibitem[Kim et~al.(2023)Kim, Lee, Kim, Ha, and Zhu]{kim2023dense}
Yunji Kim, Jiyoung Lee, Jin-Hwa Kim, Jung-Woo Ha, and Jun-Yan Zhu.
\newblock Dense text-to-image generation with attention modulation.
\newblock In \emph{Proceedings of the IEEE/CVF International Conference on
  Computer Vision}, pages 7701--7711, 2023.

\bibitem[Kumari et~al.(2022)Kumari, Zhang, Zhang, Shechtman, and
  Zhu]{kumari2022multi}
Nupur Kumari, Bingliang Zhang, Richard Zhang, Eli Shechtman, and Jun-Yan Zhu.
\newblock Multi-concept customization of text-to-image diffusion.
\newblock \emph{arXiv preprint arXiv:2212.04488}, 2022.

\bibitem[Lee et~al.(2023)Lee, Liu, Ryu, Watkins, Du, Boutilier, Abbeel,
  Ghavamzadeh, and Gu]{lee2023aligning}
Kimin Lee, Hao Liu, Moonkyung Ryu, Olivia Watkins, Yuqing Du, Craig Boutilier,
  Pieter Abbeel, Mohammad Ghavamzadeh, and Shixiang~Shane Gu.
\newblock Aligning text-to-image models using human feedback.
\newblock \emph{arXiv preprint arXiv:2302.12192}, 2023.

\bibitem[Li et~al.(2023)Li, Liu, Wu, Mu, Yang, Gao, Li, and Lee]{li2023gligen}
Yuheng Li, Haotian Liu, Qingyang Wu, Fangzhou Mu, Jianwei Yang, Jianfeng Gao,
  Chunyuan Li, and Yong~Jae Lee.
\newblock Gligen: Open-set grounded text-to-image generation.
\newblock In \emph{CVPR}, 2023.

\bibitem[Lian et~al.(2023{\natexlab{a}})Lian, Li, Yala, and
  Darrell]{lian2023llmgrounded}
Long Lian, Boyi Li, Adam Yala, and Trevor Darrell.
\newblock Llm-grounded diffusion: Enhancing prompt understanding of
  text-to-image diffusion models with large language models.
\newblock \emph{arXiv preprint arXiv:2305.13655}, 2023{\natexlab{a}}.

\bibitem[Lian et~al.(2023{\natexlab{b}})Lian, Shi, Yala, Darrell, and
  Li]{lian2023llm}
Long Lian, Baifeng Shi, Adam Yala, Trevor Darrell, and Boyi Li.
\newblock Llm-grounded video diffusion models.
\newblock \emph{arXiv preprint arXiv:2309.17444}, 2023{\natexlab{b}}.

\bibitem[Lin et~al.(2014)Lin, Maire, Belongie, Bourdev, Girshick, Hays, Perona,
  Ramanan, Doll{'{a} }r, and Zitnick]{cocodataset}
Tsung{-}Yi Lin, Michael Maire, Serge~J. Belongie, Lubomir~D. Bourdev, Ross~B.
  Girshick, James Hays, Pietro Perona, Deva Ramanan, Piotr Doll{'{a} }r, and
  C.~Lawrence Zitnick.
\newblock Microsoft {COCO:} common objects in context.
\newblock \emph{CoRR}, abs/1405.0312, 2014.

\bibitem[Liu et~al.(2022)Liu, Garrette, Saharia, Chan, Roberts, Narang, Blok,
  Mical, Norouzi, and Constant]{liu2022character}
Rosanne Liu, Dan Garrette, Chitwan Saharia, William Chan, Adam Roberts, Sharan
  Narang, Irina Blok, RJ Mical, Mohammad Norouzi, and Noah Constant.
\newblock Character-aware models improve visual text rendering.
\newblock \emph{arXiv preprint arXiv:2212.10562}, 2022.

\bibitem[Ma et~al.(2023)Ma, Lewis, Kleijn, and Leung]{ma2023directed}
Wan-Duo~Kurt Ma, JP Lewis, W~Bastiaan Kleijn, and Thomas Leung.
\newblock Directed diffusion: Direct control of object placement through
  attention guidance.
\newblock \emph{arXiv preprint arXiv:2302.13153}, 2023.

\bibitem[Nichol et~al.(2022)Nichol, Dhariwal, Ramesh, Shyam, Mishkin, McGrew,
  Sutskever, and Chen]{GLIDE}
Alex Nichol, Prafulla Dhariwal, Aditya Ramesh, Pranav Shyam, Pamela Mishkin,
  Bob McGrew, Ilya Sutskever, and Mark Chen.
\newblock Glide: Towards photorealistic image generation and editing with
  text-guided diffusion models.
\newblock 2022.

\bibitem[Nichol and Dhariwal(2022)]{nichol2021improved}
Alexander~Quinn Nichol and Prafulla Dhariwal.
\newblock Improved denoising diffusion probabilistic models.
\newblock 2022.

\bibitem[OpenAI(2023)]{openai2023gpt4}
OpenAI.
\newblock Gpt-4 technical report, 2023.

\bibitem[Paiss et~al.(2023)Paiss, Ephrat, Tov, Zada, Mosseri, Irani, and
  Dekel]{paiss2023teaching}
Roni Paiss, Ariel Ephrat, Omer Tov, Shiran Zada, Inbar Mosseri, Michal Irani,
  and Tali Dekel.
\newblock Teaching clip to count to ten, 2023.

\bibitem[Radford et~al.(2021)Radford, Kim, Hallacy, Ramesh, Goh, Agarwal,
  Sastry, Askell, Mishkin, Clark, Gretchen, and Ilya]{radford2021learning}
Alec Radford, Jong~Wook Kim, Chris Hallacy, Aditya Ramesh, Gabriel Goh,
  Sandhini Agarwal, Girish Sastry, Amanda Askell, Pamela Mishkin, Jack Clark,
  Krueger Gretchen, and Sutskever Ilya.
\newblock Learning transferable visual models from natural language
  supervision.
\newblock 2021.

\bibitem[Raffel et~al.(2020)Raffel, Shazeer, Roberts, Lee, Narang, Matena,
  Zhou, Li, and Liu]{raffel2020exploring}
Colin Raffel, Noam Shazeer, Adam Roberts, Katherine Lee, Sharan Narang, Michael
  Matena, Yanqi Zhou, Wei Li, and Peter~J Liu.
\newblock Exploring the limits of transfer learning with a unified text-to-text
  transformer.
\newblock \emph{The Journal of Machine Learning Research}, 21\penalty0
  (1):\penalty0 5485--5551, 2020.

\bibitem[Ramesh et~al.(2021)Ramesh, Pavlov, Goh, Gray, Voss, Radford, Chen, and
  Sutskever]{DALLE}
Aditya Ramesh, Mikhail Pavlov, Gabriel Goh, Scott Gray, Chelsea Voss, Alec
  Radford, Mark Chen, and Ilya Sutskever.
\newblock Zero-shot text-to-image generation.
\newblock 2021.

\bibitem[Ramesh et~al.(2022)Ramesh, Dhariwal, Nichol, Chu, and Chen]{DALLE2}
Aditya Ramesh, Prafulla Dhariwal, Alex Nichol, Casey Chu, and Mark Chen.
\newblock Hierarchical text-conditional image generation with clip latents.
\newblock \emph{arXiv preprint arXiv:2204.06125}, 2022.

\bibitem[Rombach et~al.(2022)Rombach, Blattmann, Lorenz, Esser, and
  Ommer]{rombach2022high}
Robin Rombach, Andreas Blattmann, Dominik Lorenz, Patrick Esser, and Bj{\"o}rn
  Ommer.
\newblock High-resolution image synthesis with latent diffusion models.
\newblock In \emph{Proceedings of the IEEE/CVF Conference on Computer Vision
  and Pattern Recognition}, 2022.

\bibitem[Ruiz et~al.(2023)Ruiz, Li, Jampani, Pritch, Rubinstein, and
  Aberman]{ruiz2022dreambooth}
Nataniel Ruiz, Yuanzhen Li, Varun Jampani, Yael Pritch, Michael Rubinstein, and
  Kfir Aberman.
\newblock Dreambooth: Fine tuning text-to-image diffusion models for
  subject-driven generation.
\newblock \emph{CVPR}, 2023.

\bibitem[Saharia et~al.(2022)Saharia, Chan, Saxena, Li, Whang, Denton,
  Ghasemipour, Gontijo~Lopes, Karagol~Ayan, Salimans, Jonathan, David, and
  Mohammad]{Imagen}
Chitwan Saharia, William Chan, Saurabh Saxena, Lala Li, Jay Whang, Emily~L
  Denton, Kamyar Ghasemipour, Raphael Gontijo~Lopes, Burcu Karagol~Ayan, Tim
  Salimans, Ho Jonathan, J~Fleet David, and Norouzi Mohammad.
\newblock Photorealistic text-to-image diffusion models with deep language
  understanding.
\newblock In \emph{NeurIPS}, 2022.

\bibitem[Sauer et~al.(2023)Sauer, Karras, Laine, Geiger, and
  Aila]{sauer2023stylegan}
Axel Sauer, Tero Karras, Samuli Laine, Andreas Geiger, and Timo Aila.
\newblock Stylegan-t: Unlocking the power of gans for fast large-scale
  text-to-image synthesis.
\newblock \emph{arXiv preprint arXiv:2301.09515}, 2023.

\bibitem[Schuhmann et~al.(2022)Schuhmann, Beaumont, Vencu, Gordon, Wightman,
  Cherti, Coombes, Katta, Mullis, Wortsman, Patrick, Srivatsa, Katherine,
  Ludwig, Robert, and Jenia]{schuhmann2022laion}
Christoph Schuhmann, Romain Beaumont, Richard Vencu, Cade Gordon, Ross
  Wightman, Mehdi Cherti, Theo Coombes, Aarush Katta, Clayton Mullis, Mitchell
  Wortsman, Schramowski Patrick, Kundurthy Srivatsa, Crowson Katherine, Schmidt
  Ludwig, Kaczmarczyk Robert, and Jitsev Jenia.
\newblock Laion-5b: An open large-scale dataset for training next generation
  image-text models.
\newblock \emph{NeurIPS}, 2022.

\bibitem[Song et~al.(2021)Song, Meng, and Ermon]{song2021denoising}
Jiaming Song, Chenlin Meng, and Stefano Ermon.
\newblock Denoising diffusion implicit models.
\newblock In \emph{ICLR}, 2021.

\bibitem[Touvron et~al.({\natexlab{a}})Touvron, Lavril, Izacard, Martinet,
  Lachaux, Lacroix, Rozi{\`e}re, Goyal, Hambro, Azhar,
  et~al.]{touvron2302llama}
Hugo Touvron, Thibaut Lavril, Gautier Izacard, Xavier Martinet, Marie-Anne
  Lachaux, Timoth{\'e}e Lacroix, Baptiste Rozi{\`e}re, Naman Goyal, Eric
  Hambro, Faisal Azhar, et~al.
\newblock Llama: open and efficient foundation language models, 2023.
\newblock \emph{URL https://arxiv. org/abs/2302.13971}, {\natexlab{a}}.

\bibitem[Touvron et~al.({\natexlab{b}})Touvron, Martin, Stone, Albert,
  Almahairi, Babaei, Bashlykov, Batra, Bhargava, Bhosale,
  et~al.]{touvron2307llama}
Hugo Touvron, Louis Martin, Kevin Stone, Peter Albert, Amjad Almahairi, Yasmine
  Babaei, Nikolay Bashlykov, Soumya Batra, Prajjwal Bhargava, Shruti Bhosale,
  et~al.
\newblock Llama 2: Open foundation and fine-tuned chat models. corr,
  abs/2307.09288, 2023b. doi: 10.48550.
\newblock \emph{arXiv preprint arXiv.2307.09288}, {\natexlab{b}}.

\bibitem[Touvron et~al.(2023)Touvron, Lavril, Izacard, Martinet, Lachaux,
  Lacroix, Rozière, Goyal, Hambro, Azhar, Rodriguez, Joulin, Grave, and
  Lample]{touvron2023llama}
Hugo Touvron, Thibaut Lavril, Gautier Izacard, Xavier Martinet, Marie-Anne
  Lachaux, Timothée Lacroix, Baptiste Rozière, Naman Goyal, Eric Hambro,
  Faisal Azhar, Aurelien Rodriguez, Armand Joulin, Edouard Grave, and Guillaume
  Lample.
\newblock Llama: Open and efficient foundation language models, 2023.

\bibitem[Vaswani et~al.(2017)Vaswani, Shazeer, Parmar, Uszkoreit, Jones, Gomez,
  Kaiser, and Polosukhin]{vaswani2017attention}
Ashish Vaswani, Noam Shazeer, Niki Parmar, Jakob Uszkoreit, Llion Jones,
  Aidan~N Gomez, {\L}ukasz Kaiser, and Illia Polosukhin.
\newblock Attention is all you need.
\newblock \emph{Advances in neural information processing systems}, 30, 2017.

\bibitem[Wang et~al.(2023)Wang, Bochkovskiy, and Liao]{wang2023yolov7}
Chien-Yao Wang, Alexey Bochkovskiy, and Hong-Yuan~Mark Liao.
\newblock {YOLOv7}: Trainable bag-of-freebies sets new state-of-the-art for
  real-time object detectors.
\newblock In \emph{Proceedings of the IEEE/CVF Conference on Computer Vision
  and Pattern Recognition (CVPR)}, 2023.

\bibitem[Wu et~al.(2023{\natexlab{a}})Wu, Liu, Zhao, Bui, Lin, Zhang, and
  Chang]{wu2023harnessing}
Qiucheng Wu, Yujian Liu, Handong Zhao, Trung Bui, Zhe Lin, Yang Zhang, and
  Shiyu Chang.
\newblock Harnessing the spatial-temporal attention of diffusion models for
  high-fidelity text-to-image synthesis.
\newblock \emph{arXiv preprint arXiv:2304.03869}, 2023{\natexlab{a}}.

\bibitem[Wu et~al.(2023{\natexlab{b}})Wu, Sun, Zhu, Zhao, and Li]{wu2023better}
Xiaoshi Wu, Keqiang Sun, Feng Zhu, Rui Zhao, and Hongsheng Li.
\newblock Better aligning text-to-image models with human preference.
\newblock \emph{arXiv preprint arXiv:2303.14420}, 2023{\natexlab{b}}.

\bibitem[Xiao et~al.(2023)Xiao, Yin, Freeman, Durand, and
  Han]{xiao2023fastcomposer}
Guangxuan Xiao, Tianwei Yin, William~T Freeman, Fr{\'e}do Durand, and Song Han.
\newblock Fastcomposer: Tuning-free multi-subject image generation with
  localized attention.
\newblock \emph{arXiv preprint arXiv:2305.10431}, 2023.

\bibitem[Xie et~al.(2023)Xie, Li, Huang, Liu, Zhang, Zheng, and
  Shou]{xie2023boxdiff}
Jinheng Xie, Yuexiang Li, Yawen Huang, Haozhe Liu, Wentian Zhang, Yefeng Zheng,
  and Mike~Zheng Shou.
\newblock Boxdiff: Text-to-image synthesis with training-free box-constrained
  diffusion.
\newblock \emph{arXiv preprint arXiv:2307.10816}, 2023.

\bibitem[Xue et~al.()Xue, Barua, Constant, Al-Rfou, Narang, Kale, Roberts, and
  Raffel]{xue2022byt5}
Linting Xue, Aditya Barua, Noah Constant, Rami Al-Rfou, Sharan Narang, Mihir
  Kale, Adam Roberts, and Colin Raffel.
\newblock {B}y{T}5: Towards a token-free future with pre-trained byte-to-byte
  models.
\newblock \emph{Transactions of the Association for Computational Linguistics}.

\bibitem[Yu et~al.(2022)Yu, Xu, Koh, Luong, Baid, Wang, Vasudevan, Ku, Yang,
  Ayan, Ben, Wei, Zarana, Xin, Han, Jason, and Yonghui]{Parti}
Jiahui Yu, Yuanzhong Xu, Jing~Yu Koh, Thang Luong, Gunjan Baid, Zirui Wang,
  Vijay Vasudevan, Alexander Ku, Yinfei Yang, Burcu~Karagol Ayan, Hutchinson
  Ben, Han Wei, Parekh Zarana, Li Xin, Zhang Han, Baldridge Jason, and Wu
  Yonghui.
\newblock Scaling autoregressive models for content-rich text-to-image
  generation.
\newblock \emph{Transactions on Machine Learning Research}, 2022.

\bibitem[Zhang et~al.(2023)Zhang, Rao, and Agrawala]{zhang2023adding}
Lvmin Zhang, Anyi Rao, and Maneesh Agrawala.
\newblock Adding conditional control to text-to-image diffusion models.
\newblock In \emph{Proceedings of the IEEE/CVF International Conference on
  Computer Vision}, pages 3836--3847, 2023.

\bibitem[Zhong et~al.(2023)Zhong, Huang, Wen, Qin, and Lin]{zhong2023adapter}
Shanshan Zhong, Zhongzhan Huang, Wushao Wen, Jinghui Qin, and Liang Lin.
\newblock Sur-adapter: Enhancing text-to-image pre-trained diffusion models
  with large language models.
\newblock \emph{arXiv preprint arXiv:2305.05189}, 2023.

\end{thebibliography}
}

\maketitlesupplementary

\section{Implementation details}
\label{sec:adding_set_up}
We apply Cross-Attention Refocusing and Self-Attention Refocusing losses on the attention maps of resolution $16\times 16$. 
All images are generated with $50$ steps of denoising. 
We discuss setting details for optimization during denoising steps, referring to Eq.~\supref{(7)} 
In terms of $\tau$, in the very early steps ($t=0$ or $t=1$), the cross and self-attention maps are unclear yet begin to form the layout. 
So, we just set the iteration step $\tau=2$. 
Then, to make the layout clearer ($t \in \{2,3,4\}$), $\tau$ is increased to 6 steps, which helps refine the layout if tokens do not attend to the corresponding boxes or are in the wrong boxes. 
We also apply early stopping to reduce inference time and ensure the quality of generated images. 
We observe that applying optimization in later steps can lead to quality degradation. Therefore, after the first ten denoising steps, we only update the latent when the tokens do not align with the corresponding boxes or with incorrect ones.
The initial step size $\alpha$ is set to $4$ in the first five steps, then decreases to $3$. The detail of the algorithm can be seen in Algorithm ~\ref{algorithm}

In this paper, we use a Gaussian kernel with filter size 3 × 3 and a $\sigma$ value of 0.5 for standard deviation

In terms of four baselines, layout-to-image models: layout-guidance, MultiDiffusion, Attend-and-Excite, GLIGEN,  they are set default in their original papers.

\begin{algorithm}
\DontPrintSemicolon
\caption{Denoising step with Attention-Refocusing}

\KwData{A text prompt $P$, a set of token indices $I$, each token associates with a set of bounding box $B_{i}$ , a timestep $t$, a set of iterations for refinement \{$t_1,\dots,t_k$\}, the threshold $T$, and a trained Stable Diffusion model SD.}
\KwResult{latent $x_{t-1}$ for the next timestep}

$A^t, S^t \leftarrow SD(z, P, t)$\;
$A^t \leftarrow \text{Softmax}(A_t - \text{sot}())$\;
\For {$i \in I$}{
    
    $A_i^t \leftarrow A_{t[:,:, i]}$\;
    $A_i^t \leftarrow \text{Gaussian}(A_i^t)$\;
    $L_i^{t,FG} \leftarrow 1 - \max(A_i^t \cdot \text{Mask}(B_i))$\;
    $L_i^{t,BG} \leftarrow \max(A_i^t \cdot (1- \text{Mask}(B_i)))$\;
    $L_{i,CAR} \leftarrow L_i^{FG} + L_i^{BG}$\;
    \For {$p \in \text{Mask}(B_i)$}{
       $L_p = \sum_{p \in B_i}(\text{Average}(S_p^t \cdot (1 - \text{Mask}(B_i))$ 
    }
    $L_{i,SAR} = \sum_{p}(L_p)$
}
$L_{CAR} \leftarrow \sum_i(L_{i, CAR})$\;
$L_{SAR} \leftarrow \sum_i(L_{i, SAR})$

$L \leftarrow L_{CAR} + L_{SAR}$\;
$\hat{x}_t \leftarrow x_t - \alpha_t \nabla_{x_t} L$\;
\If{$t \in \{t_1, \dots, t_k\}$}{
    \If{$L > 1 - T$}{
        $x_t \leftarrow \hat{x}_t$\;
        Go to Step 1\;
    }
}
$x_{t-1} \leftarrow SD(\hat{x}_t, P, t)$\;
\Return $x_{t-1}$\;
\label{algorithm}
\end{algorithm}


\section{Layout generation}
\label{sec:gpt-4}
Our full prompt mainly includes the three components:\\
\textbf{Instruction} specifies the task and defines the output format. This instruction helps GPT-4 perform better in layout generation tasks.\\
\textbf{In-context exemplars} are used further to enhance the model's capacity for the task. 
We supplement user prompts with multiple examples for the best context understanding. This also helps the model output the desired form of bounding boxes and their corresponding labels. \\
\textbf{User prompt} is appended to the instruction and the supporting examples. Then, the model completes the chat conversation from the user prompt and returns the layout in the defined form.

Once the user provides a prompt (user prompt), it will be added to the defined and fixed text to create a full prompt shown in ~\Tref{tab:chatgpt}. Then, the GPT-4 API completes the chat and returns the box coordinates of the corresponding objects.

\begin{figure*}[t]
    \centering
    \includegraphics[width=1.\textwidth]{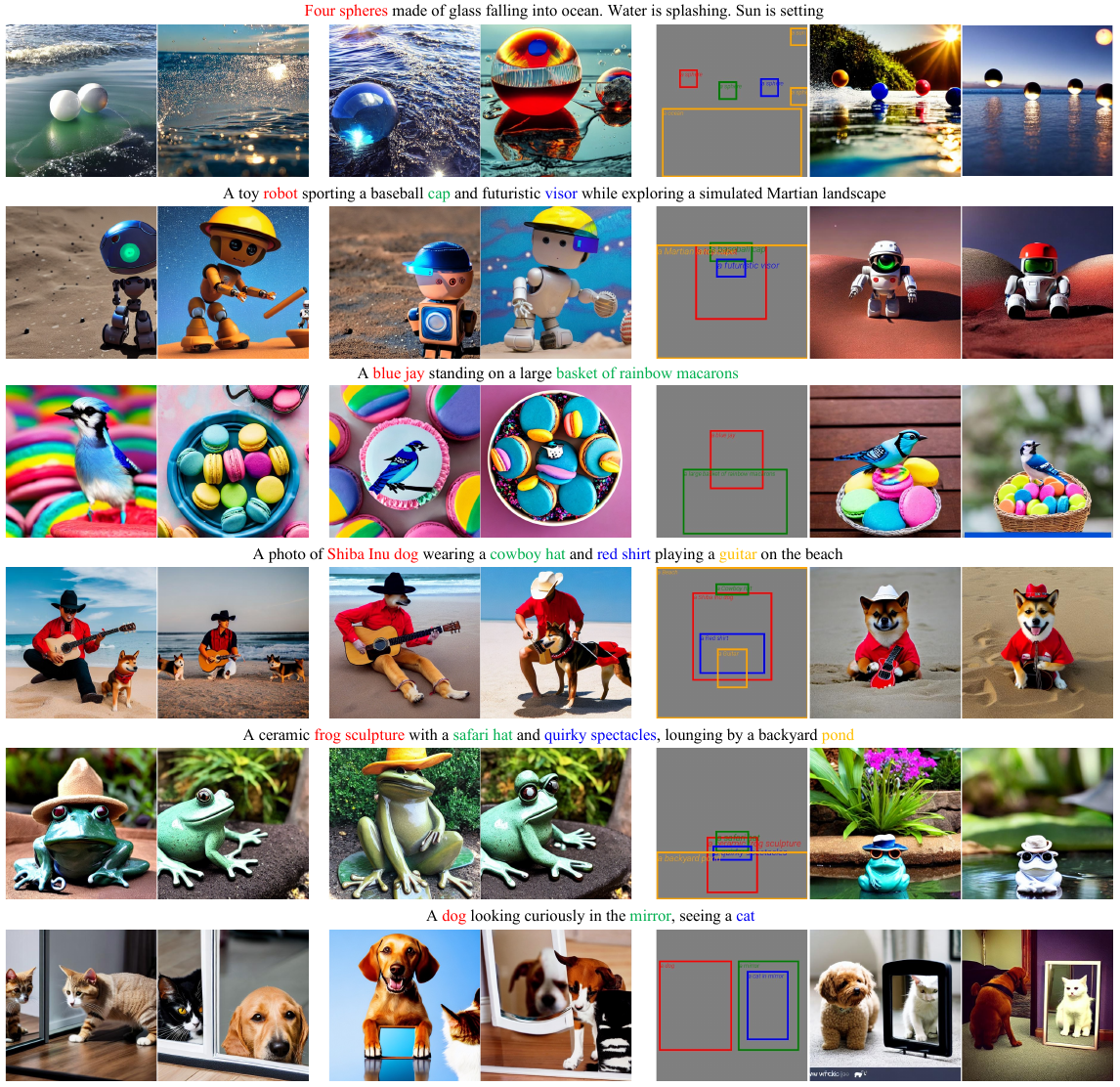}
    \begin{tabu} to 1.\textwidth {X[1.2,c]X[1.2,c]X[0.6,c]X[1.2,c]}
        \rule{0pt}{-1cm} Stable Diffusion & Attend-and-Excite & Layout from GPT-4  &  Ours
    \end{tabu}
    
    \caption{Comparison current text-to-image models (Stable diffusion and Attend-and-excite) and two-stage pipeline (layout generated from GPT-4) }
    
    \label{fig:SD_AE_ours}
\end{figure*}

The comparison of our two-stage text-to-image models with single-stage one ( Stable diffusion, Attend-and-Excite) can be seen in the  ~\Fref{fig:SD_AE_ours}

\begin{figure*}[t]
    \centering
    \includegraphics[width=1.\textwidth]{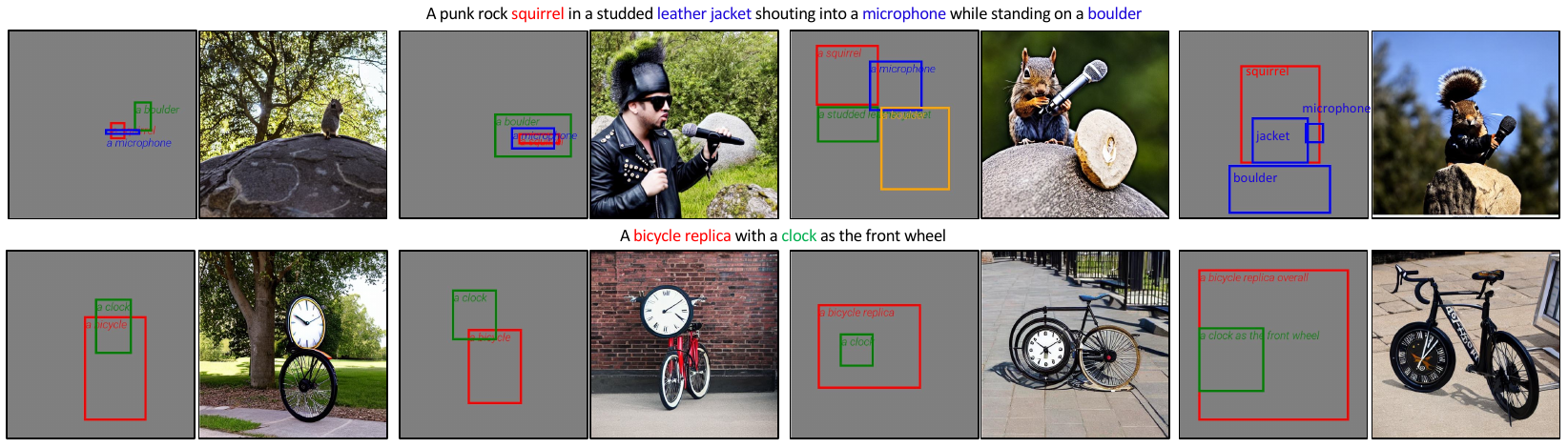}
    \begin{tabu} to 1.\textwidth {X[1.2,c]X[1.2,c]X[1.2,c]X[1.2,c]}
        \rule{0pt}{-1cm} Llama 1 ~\cite{touvron2302llama} &LLama 2 ~\cite{touvron2307llama} & GPT-3 ~\cite{brown2020language} &  GPT-4 ~\cite{openai2023gpt4}
        
    \end{tabu}
    
    \caption{Comparison generated layouts from Llama 1, Llama 2, GPT-3, GPT-4 }
    
    \label{fig:llama_gpt}
\end{figure*}

\myparagraph{Comparison of four language models}
The comparison of four language models in layout generation task is shown in ~\Fref{fig:llama_gpt}. GPT-4 is capable of reasoning implicit object relationships. For instance, in the first prompt, a squirrel with a leather racket, GPT-4 can place the leather racket box centrally within the squirrel box, unlike GPT-3, Llama 2, and Llama1, which miss the spatial composition. 
\begin{table*}
\centering
    
    \tabulinesep=4pt
    \begin{tabu} to 1.\textwidth {@{}X[1.9,l]X[9,l]@{}}
    \toprule
    \textbf{Role} & \textbf{Content} \\
    \midrule
      \textbf{Instruction}& System:{ "You are ChatGPT-4, a large language model trained by OpenAI. Your goal is to assist users by providing helpful and relevant information. In this context, you are expected to generate specific coordinate box locations for objects in a description, considering their relative sizes and positions and the number of objects. The box coordinates should be in the order ( left, top, right, bottom). The size of the image is 512*512."}\\
    \midrule
     \multirow{4}{*}{\textbf{In-context examples}} & User: {"Provide box coordinates for an image with a cat in the middle of a car and a chair. Make the size of the boxes as big as possible."}\\
     & Assistant: "cat: (245, 176, 345, 336); car: (10, 128, 230, 384);  chair: (353, 224, 498, 350)" \\
    & User : "Provide box coordinates for an image with three cats on the field."\\
    
     &Assistant: "cat: (51, 82, 399, 279);cat: (288, 128, 472, 299); cat: (27, 355, 418, 494)"\\
    \midrule
    \textbf{User prompt} & User : "Provide the Provide box coordinates for an image with" + [user prompt]\\
    \bottomrule
    \end{tabu}
    \caption{The full prompt for gpt4 api.  }
    \label{tab:chatgpt}
\end{table*}

    
    

    
    

\begin{table}[h]
    \centering
    
    \small{
    \tabulinesep=0pt
    \begin{tabu} to 1.\columnwidth {@{}X[0.6,c]X[0.6,c]X[0.6,c]X[1.1,c]X[0.6,c]X[1,c]X[1,c]@{}}
    \toprule
    \multicolumn{5}{c}{Stable Diffusion } & \multicolumn{2}{c}{GLIGEN}\\
    \cmidrule(r){1-5} \cmidrule(r){6-7}
    {SD} & {+AE} & {+MD} & {+LG} & {+Ours} & {GLIGEN} & {+Ours}\\ 
    \midrule
    54.33 & 101.67 & 74.16 & 111.13 & 102.97 & 205.90 & 279.08\\ 
    \bottomrule
    \end{tabu}}
    \caption{Inference time of different methods (s$/$10 images). AE: Attend-and-Excite, MD: MultiDiffusion, LG: Layout-guidance, Ours: Attention-Refocusing}
    \label{tab:time}
\end{table}

\section{Applying CAR loss for segmentation mask}
\label{sec:adding_seg}
\begin{figure*}[t]

    \centering
    \includegraphics[width=1\textwidth]{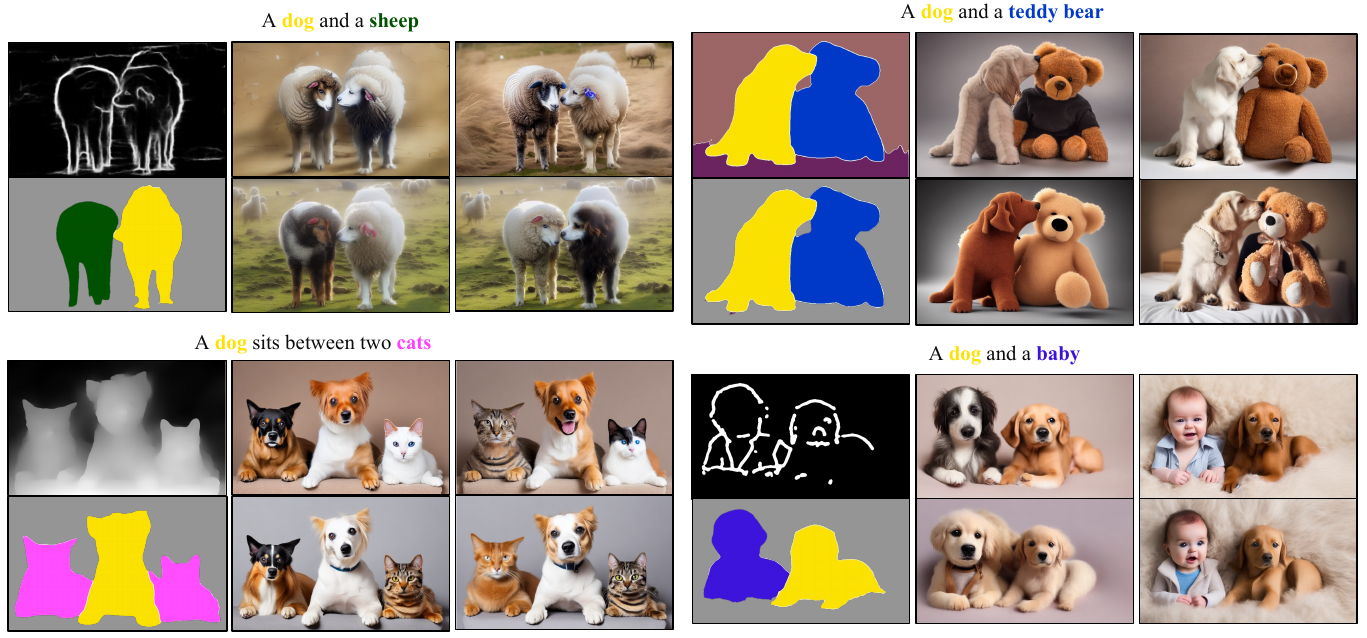}
  
      \begin{tabu} to 1.\textwidth {X[2cm,c]X[1.9cm,c]X[1.9cm,c]X[1.9cm,c]X[1.9cm,c]X[1.9cm,c]}
        Input & ControlNet ~\cite{zhang2023adding}&ControlNet + Ours & Input & ControlNet ~\cite{zhang2023adding} & ControlNet + Ours\\
       
        \end{tabu}
    
\caption{
\tb{ControlNet with attention-based guidance.}
In the input columns, upper representations are inputs of ControlNet, and lower ones are segmentation masks used for our losses. Applying our losses can refine the attribute blending of ControlNet.}
\label{fig:controlnet_append}
\end{figure*}

We also adapt the CAR loss to other layout modalities like depth maps, segmentation masks, and edge maps. 
Specifically, we always use the converted segmentation masks $M_i$ associated with token $i$-th to apply our method. 
Since the segmentation provides a precise object boundary in contrast to the bounding box, we optimize the attention over the entire foreground by taking the average instead of the maximum. 
The foreground loss for segmentation masks is: 
\begin{equation}
  \mathcal{L}_{FG} = \frac{1}{q} \sum_{i \in I}  \frac{\sum(1 - (A_i^t \cdot  M_i))}{\sum M_i}
\end{equation}
Similarly, the background loss for segmentation maps is:
\begin{equation}
  \mathcal{L}_{BG} = \frac{1}{q} \sum_{i \in I} \frac{\sum A_i^t \cdot (1- M_i)}{\sum (1 - M_i)}
\end{equation}
The $L_{SAR}$ is calculated using the formulation for bounding box presented in the main paper.
The more results of applying our losses to ControlNet ~\cite{zhang2023adding} are shown in ~\Fref{fig:controlnet_append}
\section{Additional quantitative result}
\label{sec:additional_quant}
The ~\Tref{tab:drawbench} shows results of our methods in drawbench ~\cite{Imagen}. Our proposed losses demonstrate a comparable performance boost to HRS. 
By integrating our losses, we compare favorably or comparatively against baselines in the counting procedure. 
Moreover, our losses substantially improve the accuracy of the spatial category. 

~\Tref{tab:boxdiff} compares our attention-refocusing method and Boxdiff ~\cite{xie2023boxdiff}. It can be seen that Attention-refocusing losses outperform Boxdiff in counting, spatial and color categories, especially in spatial composition, our method surpasses Boxdiff around 7\%. 

 ~\Tref{tab:counting_hrs} shows the full quantitative result in counting in the HRS benchmark.
We we compare the time inference of our losses and other free-training methods in  ~\Tref{tab:time}. Our losses are compatible with Attend-and-excite ~\cite{chefer2023attend} , even more effective than Layout-guidance ~\cite{chen2023training} in speed.
\begin{table}[t]
    \centering
    
    \footnotesize{
    \tabulinesep=0pt
    \begin{tabu} to 1.\columnwidth {@{}X[1.7, l]X[0.7,c]X[1.3,c]X[1.1,c]X[1.3,c]X[1.2,c]@{}}
        \toprule
        \multirow{2}{*}{\textbf{Method}} & \multirow{2}{*}{\shortstack[c]{\textbf{CAR} \\ \& \textbf{SAR}}} &\multicolumn{3}{c}{\textbf{Counting}} &\hspace{-0.3cm} \textbf{Spatial}\\
         \cmidrule(r){3-5} \cmidrule(r){6-6}
         \footnotesize & & Precision~$\uparrow$ & \hspace{-0.2cm}Recall~$\uparrow$ & \hspace{-0.45cm}F1~$\uparrow$& \hspace{-0.3cm}Accuracy~$\uparrow$ \\
        \midrule
        \multirow{2}{*}{\shortstack[l]{Stable Diffusion \\ \cite{rombach2022high}}} & \xmark &  \hspace{-0.4cm}73.32   & \hspace{-0.35cm}70.00  &  \hspace{-0.45cm}71.55  &  \hspace{-0.4cm}12.50 \\
        & \cmark & 78.53~\goodi{5.2}   & 73.63~\goodi{3.6}  &  75.81~\goodi{4.3} &  43.50~\goodi{31.0} \\
        \midrule
        \multirow{2}{*}{\shortstack[l]{Attend-and-excite \\ \cite{chefer2023attend}}} & \xmark &  \hspace{-0.4cm}77.64   & \hspace{-0.35cm}74.85  &  \hspace{-0.45cm}76.20 &  \hspace{-0.4cm}20.50 \\
        & \cmark & 74.06~\badi{3.6}  & 77.58~\goodi{2.7}  &  75.66~\badi{0.5} &  38.00~\goodi{18.0} \\
        \midrule
        \multirow{2}{*}{\shortstack[l]{Layout-guidance \\ \cite{chen2023training}}} & \xmark & \hspace{-0.4cm}79.15    & \hspace{-0.35cm}70.61  &  \hspace{-0.45cm}74.48 & \hspace{-0.4cm}36.50   \\
        & \cmark & 78.45~\badi{0.7} &  75.45~\goodi{4.8} & 76.82~\goodi{2.3}  &52.50~\goodi{16.0}\\
        \midrule
        \multirow{2}{*}{\shortstack[l]{MultiDiffusion \\ \cite{bar2023multidiffusion}}} & \xmark & \hspace{-0.4cm}75.37    &  \hspace{-0.35cm}65.61 & \hspace{-0.45cm}69.90  &  \hspace{-0.4cm}38.00  \\
        &\cmark & 84.30~\goodi{8.9} & 68.03~\goodi{2.4} & 75.20~\goodi{5.3} & 54.50~\goodi{16.5} \\
        \midrule
        \multirow{2}{*}{\shortstack[l]{GLIGEN \\ \cite{li2023gligen}}} & \xmark & \hspace{-0.4cm}81.66 &  \hspace{-0.35cm}80.89 &  \hspace{-0.45cm}81.18 & \hspace{-0.4cm}48.00   \\
        & \cmark & 90.28~\goodi{8.6} & 86.21~\goodi{5.3} & 88.16~\goodi{7.0} & 64.00~\goodi{16.0}\\
        \bottomrule
    \end{tabu}}
    \caption{Quantitavie evaluation on the DrawBench benchmark.}
    \label{tab:drawbench}
    \vspace{-1em}
\end{table}

\begin{table}[t!]
    \centering

    \small{
    \tabulinesep=0pt
    \begin{tabu} to 1.\columnwidth {@{}X[2.2, m]X[0.7,c]X[1.4,c]X[1.4,c]X[1.3,c]@{}}
        \toprule
        \vspace{-1em}\textbf{Method} & \shortstack[c]{\textbf{CAR} \\ \& \textbf{SAR}} & \vspace{-1.2em}\textbf{Precision~$\uparrow$} & \vspace{-1.2em}\hspace{-0.5cm}\textbf{Recall~$\uparrow$} & \vspace{-1.2em}\hspace{-0.5cm}\textbf{F1~$\uparrow$}\\
        \midrule
        \multirow{2}{*}{\shortstack[l]{Stable Diffusion \cite{rombach2022high}}} & \xmark & \hspace{-0.6cm}71.86 & \hspace{-0.6cm}52.19 & \hspace{-0.58cm}58.31 \\
        & \cmark & 81.56~\good{9.7}& 51.19~\bad{1.0} & 60.62~\good{2.3} \\
        \midrule
        \multirow{2}{*}{\shortstack[l]{Attend-and-excite \cite{chefer2023attend}}} & \xmark &  \hspace{-0.6cm}73.10 & \hspace{-0.6cm}54.79 & \hspace{-0.58cm}60.47 \\
        & \cmark & 75.94~\good{2.8} & 56.31~\good{1.5} & 62.71~\good{2.2} \\
        \midrule
        \multirow{2}{*}{\shortstack[l]{Layout-guidance \cite{chen2023training}}} & \xmark & \hspace{-0.6cm}80.60 & \hspace{-0.6cm}45.83 & \hspace{-0.58cm}56.22   \\
        & \cmark &78.15~\bad{2.5}& 55.65~\good{9.8} & 63.01~\good{6.8} \\
        \midrule
        \multirow{2}{*}{\shortstack[l]{MultiDiffusion\cite{bar2023multidiffusion}  }} & \xmark & \hspace{-0.6cm}78.96 & \hspace{-0.6cm}45.18 & \hspace{-0.58cm}55.18  \\
        &\cmark & 83.26~\good{4.3}& 45.71~\good{0.5} & 57.37~\good{2.2} \\
        \midrule
        \multirow{2}{*}{\shortstack[l]{GLIGEN \cite{li2023gligen} }} & \xmark & \hspace{-0.6cm}78.81 & \hspace{-0.6cm}59.44 & \hspace{-0.58cm}66.58  \\
        & \cmark & 81.25~\good{2.4}&59.39~\badi{-0.1} & 67.54~\good{0.7} \\
        \midrule
        Boxdiff \cite{xie2023boxdiff} & - &\hspace{-0.6cm} 83.78 & \hspace{-0.6cm} 57.81 & \hspace{-0.58cm} 67.02\\
        \bottomrule
    \end{tabu}}
    \caption{Our proposed losses improve the baselines in the HRS Counting benchmark.}
    \label{tab:counting_hrs}
\end{table}

\section{Additional visual comparison GLIGEN with and without our losses}
\label{sec:additional_plug}
\begin{figure*}[t]
    \vspace{-0.05em}
     \begin{subfigure}[t]{0.02\textwidth}
        \vspace{-8.2cm}
        \rotatebox[origin=c]{90}{
            
            $\overbrace{\hspace{4.cm}}_{\substack{\vspace{-7.0mm}\\\colorbox{white}{~~GLIGEN + Ours~~}}}$
            $\overbrace{\hspace{4.cm}}_{\substack{\vspace{-7.0mm}\\\colorbox{white}{~~GLIGEN~~}}}$            
        }
    \end{subfigure}
    \begin{subfigure}[b]{0.97\textwidth}
        \centering
        \includegraphics[width=1.\textwidth]{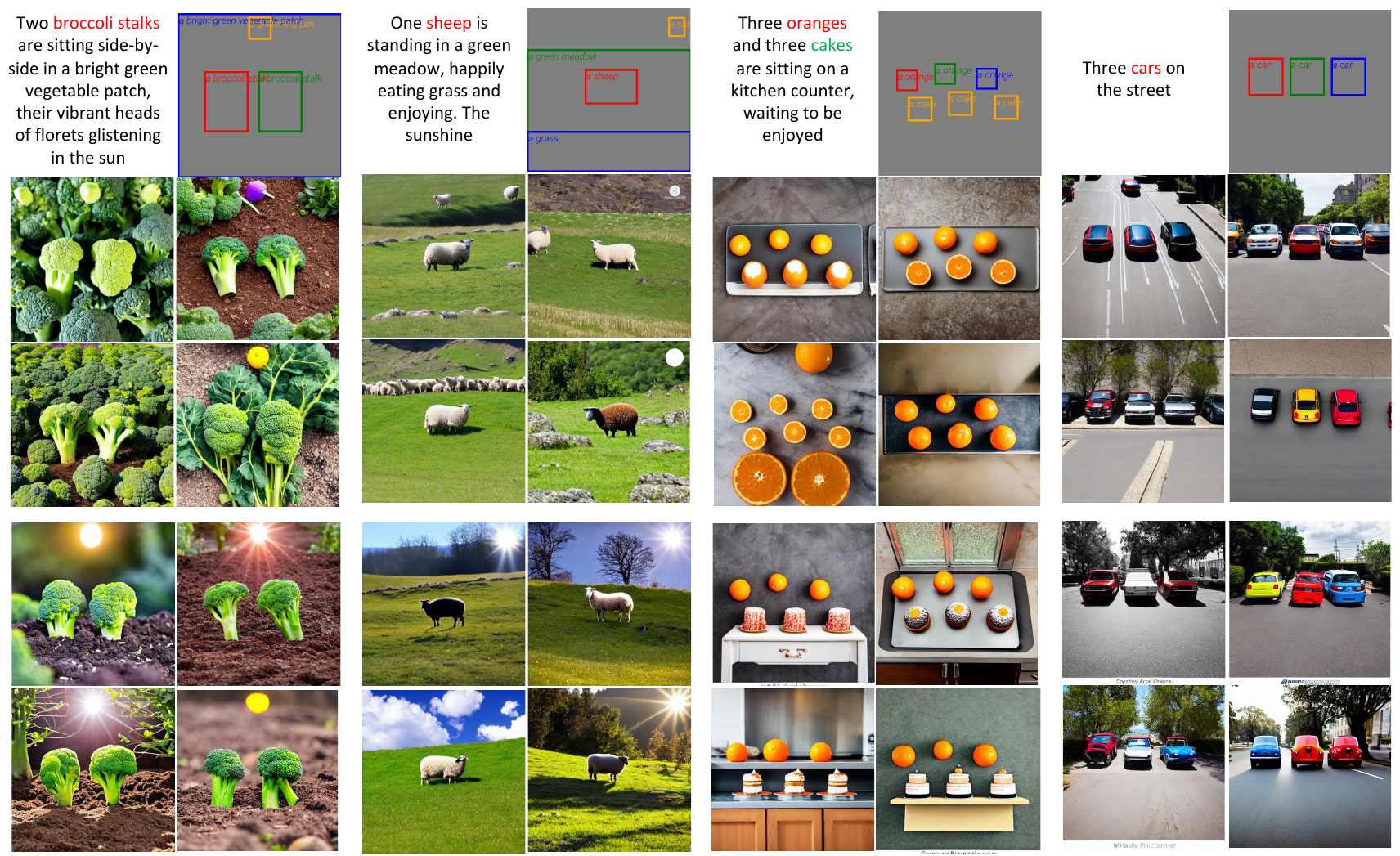}
        \vskip -0.1cm

    \end{subfigure}
    \vspace{-0.2cm}
    \caption{
    \tb{Visual comparisons of GLIGEN and GLIGEN + Ours in counting.}
    Here we apply our attention-based guidance on GLIGEN. The CAR \& SAR losses avoid extra objects in the background, leading to generating correct number of objects.}
    \label{fig:gligen_our_counting}
    \vspace{-1em}
\end{figure*}

\begin{figure*}[t]
    \vspace{-0.05em}
     \begin{subfigure}[t]{0.02\textwidth}
        \vspace{-8.2cm}
        \rotatebox[origin=c]{90}{
            
            $\overbrace{\hspace{4.cm}}_{\substack{\vspace{-7.0mm}\\\colorbox{white}{~~GLIGEN + Ours~~}}}$
            $\overbrace{\hspace{4.cm}}_{\substack{\vspace{-7.0mm}\\\colorbox{white}{~~GLIGEN~~}}}$            
        }
    \end{subfigure}
    \begin{subfigure}[b]{0.97\textwidth}
        \centering
        \includegraphics[width=1.\textwidth]{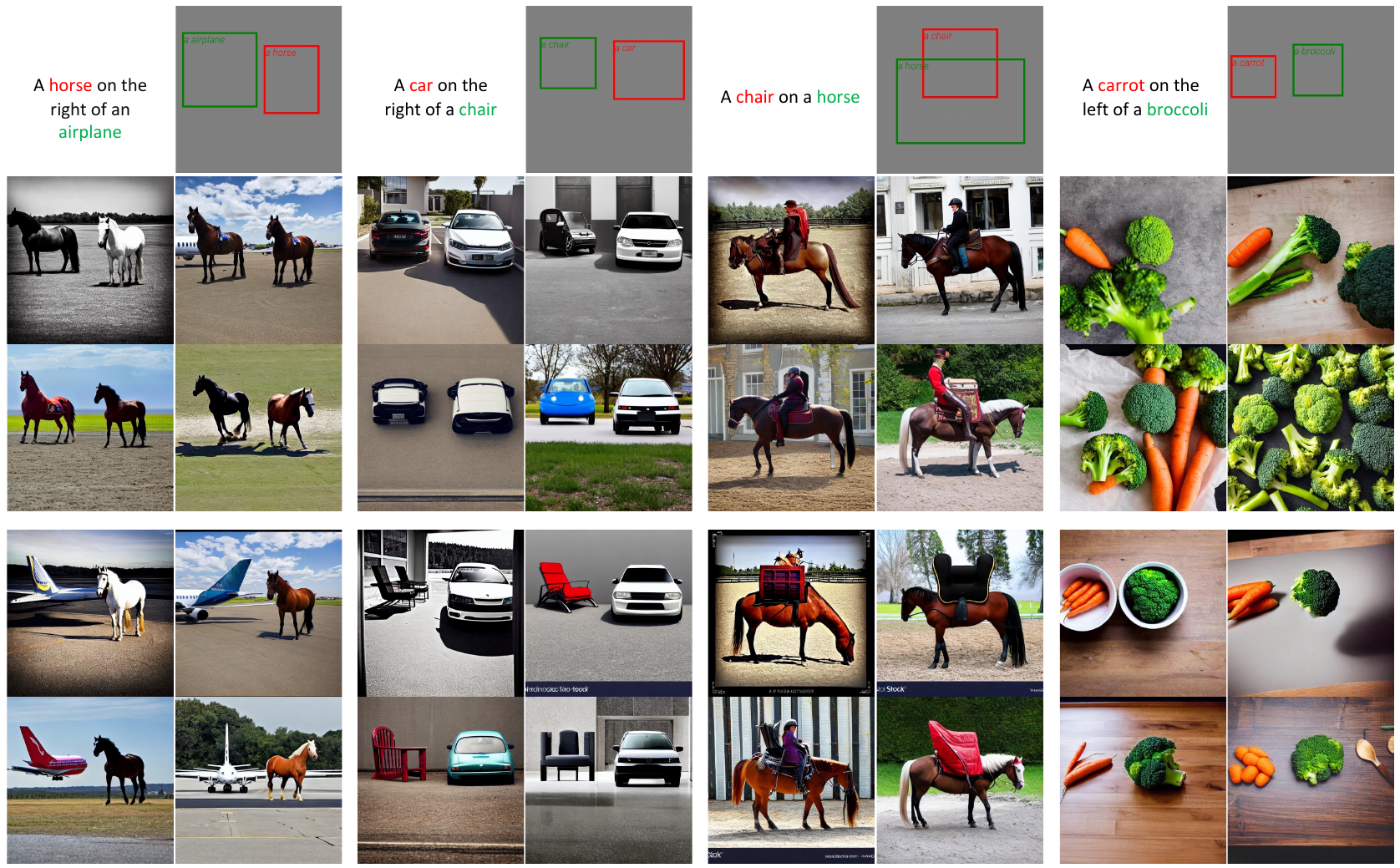}
        \vskip -0.1cm

    \end{subfigure}
    \vspace{-0.2cm}
    \caption{\tb{Visual comparisons of GLIGEN and GLIGEN + Ours in spatial.}The CAR \& SAR losses mitigate the problem that objects are generated in incorrect boxes. }

    \label{fig:gligen_our_spatial}
    \vspace{-1em}
\end{figure*}

\begin{figure*}[t]
     \begin{subfigure}[t]{0.02\textwidth}
        \vspace{-8.2cm}
        \rotatebox[origin=c]{90}{
            
            $\overbrace{\hspace{4.cm}}_{\substack{\vspace{-7.0mm}\\\colorbox{white}{~~GLIGEN + Ours~~}}}$
            $\overbrace{\hspace{4.cm}}_{\substack{\vspace{-7.0mm}\\\colorbox{white}{~~GLIGEN~~}}}$            
        }
    \end{subfigure}
    \begin{subfigure}[b]{0.97\textwidth}
        \centering
        \includegraphics[width=1.\textwidth]{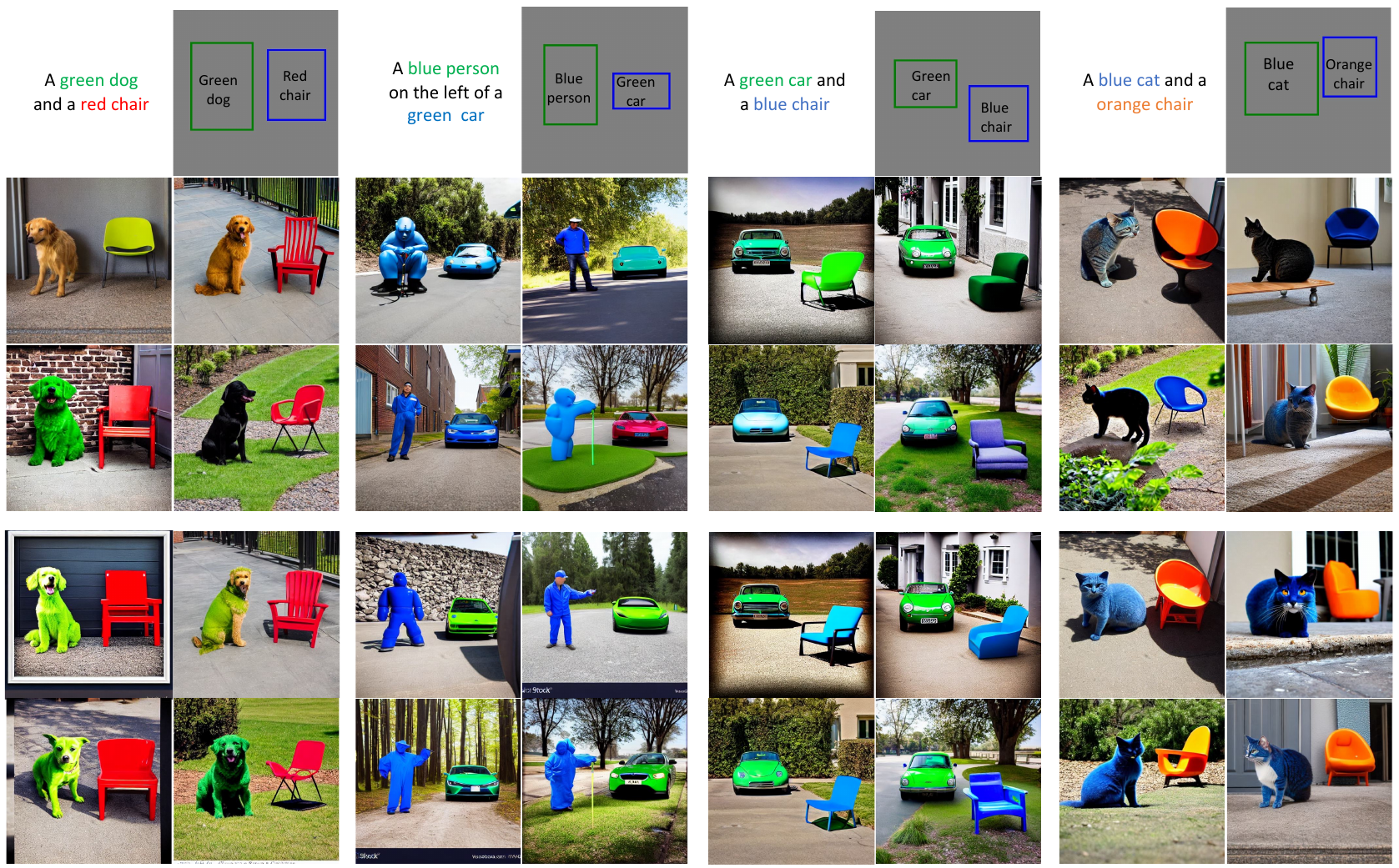}
        \vskip -0.1cm

    \end{subfigure}
    \vspace{-0.2cm}
    \caption{
    \tb{Visual comparisons of GLIGEN and GLIGEN + Ours in color. } With Attention-Refocusing losses, we can refine color blinding problem in GLIGEN.
    }
    \label{fig:gligen_our_color}
    \vspace{-1em}
\end{figure*}

\begin{figure*}[t]
    \vspace{-0.05em}
     \begin{subfigure}[t]{0.02\textwidth}
        \vspace{-8.2cm}
        \rotatebox[origin=c]{90}{
            
            $\overbrace{\hspace{4.cm}}_{\substack{\vspace{-7.0mm}\\\colorbox{white}{~~GLIGEN + Ours~~}}}$
            $\overbrace{\hspace{4.cm}}_{\substack{\vspace{-7.0mm}\\\colorbox{white}{~~GLIGEN~~}}}$            
        }
    \end{subfigure}
    \begin{subfigure}[b]{0.97\textwidth}
        \centering
        \includegraphics[width=1.\textwidth]{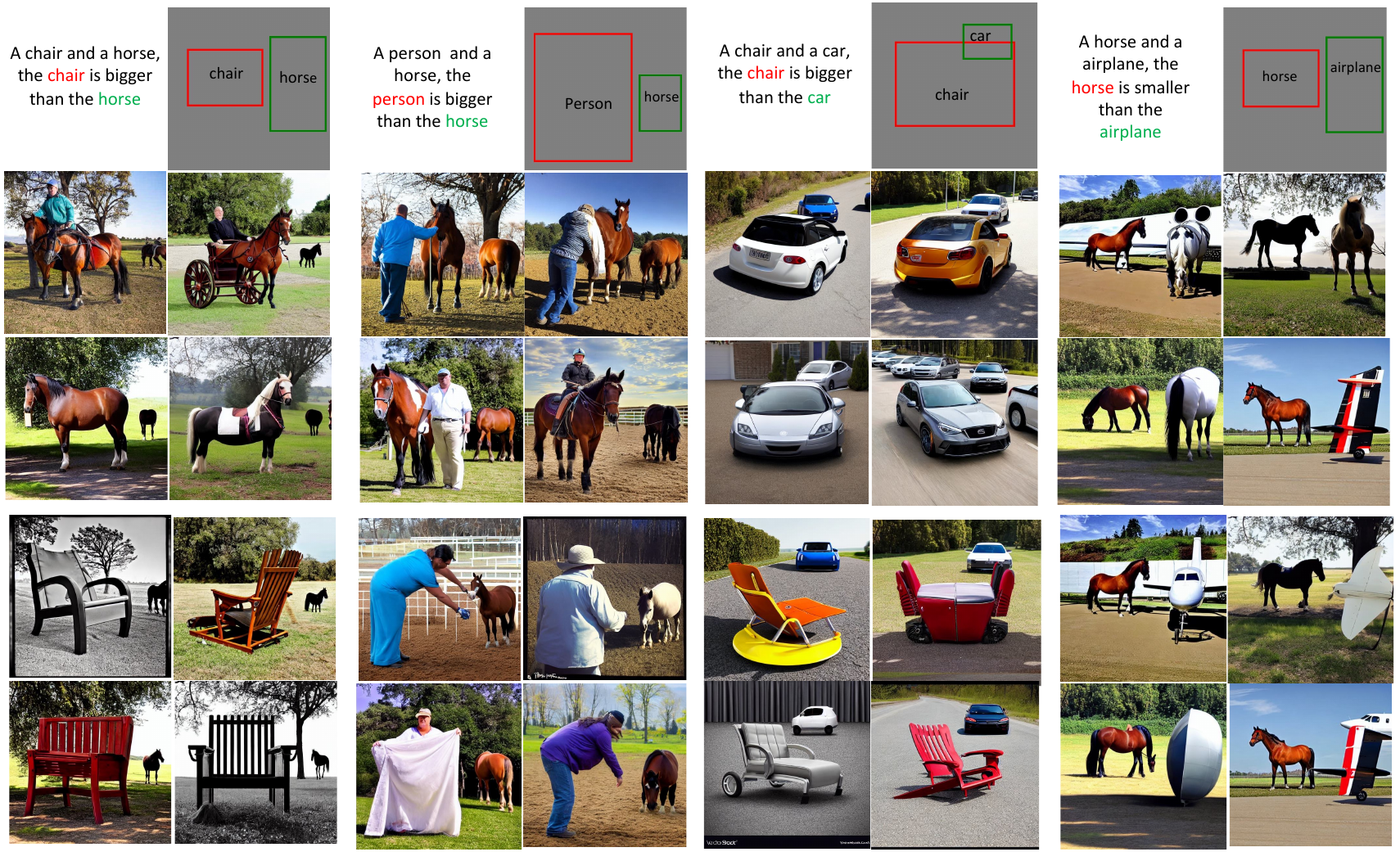}
        \vskip -0.1cm

    \end{subfigure}
    \vspace{-0.1cm}
    \caption{
    \tb{Visual comparisons of GLIGEN and GLIGEN + Ours in size.} Applying our losses can generate correct objects in the bounding boxes, leading to improving the accuracy in size category ( even with out-of- distribution prompts)}
    \label{fig:gligen_our_size}
    \vspace{-1em}
\end{figure*}

The ~\Fref{fig:gligen_our_counting},~\Fref{fig:gligen_our_spatial}, ~\Fref{fig:gligen_our_color}, and ~\Fref{fig:gligen_our_size} show more comparions in three categories: counting, spatial, size and color compositions. Our attention-refocusing losses effectively refine misaligned objects in GLIGEN, the strongest baseline. 

\begin{figure*}[h]
    \vspace{-0.5em}
     \begin{subfigure}[t]{0.02\textwidth}

        \vspace{-16.8cm}
        \rotatebox[origin=c]{90}{
            $\overbrace{\hspace{4cm}}_{\substack{\vspace{-6.8mm}\\\colorbox{white}{~~GLIGEN+ Ours~~}}}$
            $\overbrace{\hspace{4.cm}}_{\substack{\vspace{-7.0mm}\\\colorbox{white}{~~GLIGEN~\cite{li2023gligen}~~}}}$
            $\overbrace{\hspace{4. cm}}_{\substack{\vspace{-7.mm}\\\colorbox{white}{~~MultiDiffusion~\cite{bar2023multidiffusion} ~~}}}$
            $\overbrace{\hspace{4.3cm}}_{\substack{\vspace{-7.0mm}\\\colorbox{white}{~~Layout-guidance~\cite{chen2023training}~~}}}$            
        }
        
    \end{subfigure}
    \begin{subfigure}[b]{0.97\textwidth}
        \centering
        \includegraphics[width=1.\textwidth]{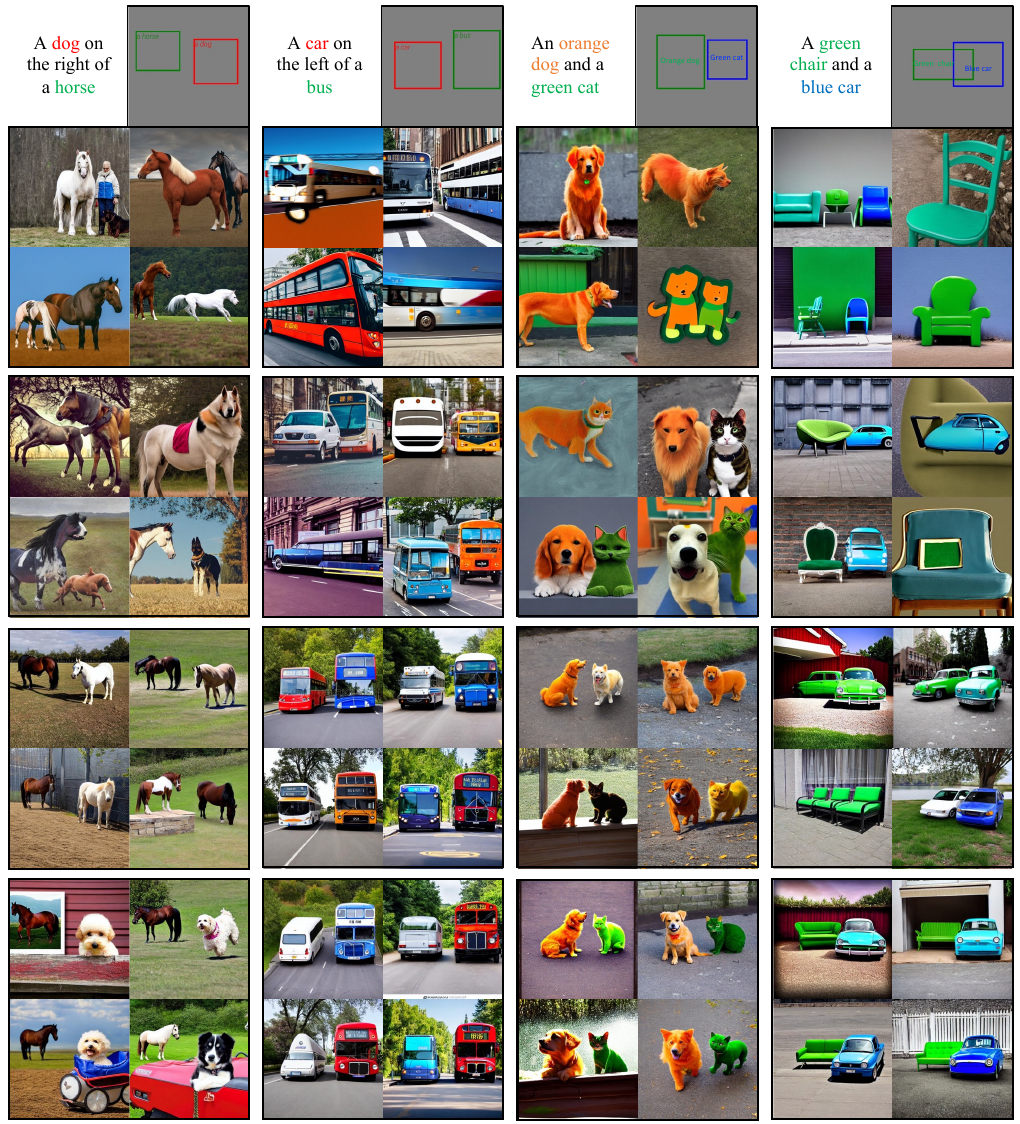}
        \vskip -0.1cm
       
    \end{subfigure}
    \caption{
    \tb{Visual comparisons on HRS and Drawbench benchmark.}
    Here we apply our attention-based guidance on grouned text-to-image models.
    All methods take the same grounded texts as inputs.
    The results show the capability of our method in synthesizing novel spatial compositions and attributes.
    }
    \label{fig:additional_baselines}
\end{figure*}

We also provide additional results to compare four baselines grounded text-to-image models in  ~\Fref{fig:additional_baselines} for HRS and Drawbench. It can be seen that our losses build upon GLIGEN is outperform other baselines, text-to-image models in terms of the four categories. 

\begin{table}[t!]
    \centering
    
    \footnotesize{
    \tabulinesep=0pt
    \begin{tabu} to 1.\columnwidth {@{}X[2.6, l]X[1.3,c]X[1.3,c]X[1.3,c]X[1.3,c]@{}}
        \toprule
        \multirow{2}{*}{\textbf{Method}} &\textbf{Counting}&\multicolumn{3}{c}{ \textbf{Compositions}}\\
         \cmidrule(r){2-2} \cmidrule(r){3-5}
         \footnotesize &\hspace{-0.2cm} F1~$\uparrow$ & \hspace{-0.2cm}Spatial~$\uparrow$ & \hspace{-0.45cm}Size~$\uparrow$& \hspace{-0.3cm}Color~$\uparrow$ \\
        \midrule
        GLIGEN + ours & \hspace{-0.38cm}\textbf{67.54} & \hspace{-0.5cm}\textbf{40.22} & \hspace{-0.5cm}27.74 & \hspace{-0.6cm} \textbf{26.32} \\
        \midrule
        GLIGEN + Boxdiff  & \hspace{-0.38cm} 67.02 &\hspace{-0.5cm} 33.93 & \hspace{-0.5cm}  \textbf{28.54 }&\hspace{-0.6cm} 22.50 \\
        \bottomrule
    \end{tabu}
    }
    \caption{Comparison of Attention-refocusing and Boxdiff in the F1 score in counting and  accuracy(\%) in all spatial, size, and color categories in the HRS benchmark.}
    \label{tab:boxdiff}
\end{table}


\end{document}